\documentclass[10pt, a4paper, logo, twocolumn, copyright]{googledeepmind}

\pdftrailerid{redacted}

\makeatletter
\renewcommand\bibentry[1]{\nocite{#1}{\frenchspacing\@nameuse{BR@r@#1\@extra@b@citeb}}}
\makeatother

\usepackage{kantlipsum, lipsum}
\usepackage{dsfont}
\usepackage{dm-colors}

\usepackage[numbers]{natbib}

\usepackage{amsmath}
\usepackage{amssymb}
\usepackage{array}
\usepackage{multirow}
\usepackage{fancyvrb}
\usepackage{color}
\usepackage[utf8]{inputenc}
\makeatletter
\def\PY@reset{\let\PY@it=\relax \let\PY@bf=\relax \let\PY@ul=\relax \let\PY@tc=\relax \let\PY@bc=\relax \let\PY@ff=\relax}
\def\PY@tok#1{\csname PY@tok@#1\endcsname}
\def\PY@toks#1+{\ifx\relax#1\empty\else \PY@tok{#1}\expandafter\PY@toks\fi}
\def\PY@do#1{\PY@bc{\PY@tc{\PY@ul{\PY@it{\PY@bf{\PY@ff{#1}}}}}}}
\def\PY#1#2{\PY@reset\PY@toks#1+\relax+\PY@do{#2}}

\expandafter\def\csname PY@tok@w\endcsname{\def\PY@tc##1{\textcolor[rgb]{0.73,0.73,0.73}{##1}}}
\expandafter\def\csname PY@tok@c\endcsname{\let\PY@it=\textit\def\PY@tc##1{\textcolor[rgb]{0.25,0.50,0.50}{##1}}}
\expandafter\def\csname PY@tok@cp\endcsname{\def\PY@tc##1{\textcolor[rgb]{0.74,0.48,0.00}{##1}}}
\expandafter\def\csname PY@tok@k\endcsname{\let\PY@bf=\textbf\def\PY@tc##1{\textcolor[rgb]{0.00,0.50,0.00}{##1}}}
\expandafter\def\csname PY@tok@kp\endcsname{\def\PY@tc##1{\textcolor[rgb]{0.00,0.50,0.00}{##1}}}
\expandafter\def\csname PY@tok@kt\endcsname{\def\PY@tc##1{\textcolor[rgb]{0.69,0.00,0.25}{##1}}}
\expandafter\def\csname PY@tok@o\endcsname{\def\PY@tc##1{\textcolor[rgb]{0.40,0.40,0.40}{##1}}}
\expandafter\def\csname PY@tok@ow\endcsname{\let\PY@bf=\textbf\def\PY@tc##1{\textcolor[rgb]{0.67,0.13,1.00}{##1}}}
\expandafter\def\csname PY@tok@nb\endcsname{\def\PY@tc##1{\textcolor[rgb]{0.00,0.50,0.00}{##1}}}
\expandafter\def\csname PY@tok@nf\endcsname{\def\PY@tc##1{\textcolor[rgb]{0.00,0.00,1.00}{##1}}}
\expandafter\def\csname PY@tok@nc\endcsname{\let\PY@bf=\textbf\def\PY@tc##1{\textcolor[rgb]{0.00,0.00,1.00}{##1}}}
\expandafter\def\csname PY@tok@nn\endcsname{\let\PY@bf=\textbf\def\PY@tc##1{\textcolor[rgb]{0.00,0.00,1.00}{##1}}}
\expandafter\def\csname PY@tok@ne\endcsname{\let\PY@bf=\textbf\def\PY@tc##1{\textcolor[rgb]{0.82,0.25,0.23}{##1}}}
\expandafter\def\csname PY@tok@nv\endcsname{\def\PY@tc##1{\textcolor[rgb]{0.10,0.09,0.49}{##1}}}
\expandafter\def\csname PY@tok@no\endcsname{\def\PY@tc##1{\textcolor[rgb]{0.53,0.00,0.00}{##1}}}
\expandafter\def\csname PY@tok@nl\endcsname{\def\PY@tc##1{\textcolor[rgb]{0.63,0.63,0.00}{##1}}}
\expandafter\def\csname PY@tok@ni\endcsname{\let\PY@bf=\textbf\def\PY@tc##1{\textcolor[rgb]{0.60,0.60,0.60}{##1}}}
\expandafter\def\csname PY@tok@na\endcsname{\def\PY@tc##1{\textcolor[rgb]{0.49,0.56,0.16}{##1}}}
\expandafter\def\csname PY@tok@nt\endcsname{\let\PY@bf=\textbf\def\PY@tc##1{\textcolor[rgb]{0.00,0.50,0.00}{##1}}}
\expandafter\def\csname PY@tok@nd\endcsname{\def\PY@tc##1{\textcolor[rgb]{0.67,0.13,1.00}{##1}}}
\expandafter\def\csname PY@tok@s\endcsname{\def\PY@tc##1{\textcolor[rgb]{0.73,0.13,0.13}{##1}}}
\expandafter\def\csname PY@tok@sd\endcsname{\let\PY@it=\textit\def\PY@tc##1{\textcolor[rgb]{0.73,0.13,0.13}{##1}}}
\expandafter\def\csname PY@tok@si\endcsname{\let\PY@bf=\textbf\def\PY@tc##1{\textcolor[rgb]{0.73,0.40,0.53}{##1}}}
\expandafter\def\csname PY@tok@se\endcsname{\let\PY@bf=\textbf\def\PY@tc##1{\textcolor[rgb]{0.73,0.40,0.13}{##1}}}
\expandafter\def\csname PY@tok@sr\endcsname{\def\PY@tc##1{\textcolor[rgb]{0.73,0.40,0.53}{##1}}}
\expandafter\def\csname PY@tok@ss\endcsname{\def\PY@tc##1{\textcolor[rgb]{0.10,0.09,0.49}{##1}}}
\expandafter\def\csname PY@tok@sx\endcsname{\def\PY@tc##1{\textcolor[rgb]{0.00,0.50,0.00}{##1}}}
\expandafter\def\csname PY@tok@m\endcsname{\def\PY@tc##1{\textcolor[rgb]{0.40,0.40,0.40}{##1}}}
\expandafter\def\csname PY@tok@gh\endcsname{\let\PY@bf=\textbf\def\PY@tc##1{\textcolor[rgb]{0.00,0.00,0.50}{##1}}}
\expandafter\def\csname PY@tok@gu\endcsname{\let\PY@bf=\textbf\def\PY@tc##1{\textcolor[rgb]{0.50,0.00,0.50}{##1}}}
\expandafter\def\csname PY@tok@gd\endcsname{\def\PY@tc##1{\textcolor[rgb]{0.63,0.00,0.00}{##1}}}
\expandafter\def\csname PY@tok@gi\endcsname{\def\PY@tc##1{\textcolor[rgb]{0.00,0.63,0.00}{##1}}}
\expandafter\def\csname PY@tok@gr\endcsname{\def\PY@tc##1{\textcolor[rgb]{1.00,0.00,0.00}{##1}}}
\expandafter\def\csname PY@tok@ge\endcsname{\let\PY@it=\textit}
\expandafter\def\csname PY@tok@gs\endcsname{\let\PY@bf=\textbf}
\expandafter\def\csname PY@tok@gp\endcsname{\let\PY@bf=\textbf\def\PY@tc##1{\textcolor[rgb]{0.00,0.00,0.50}{##1}}}
\expandafter\def\csname PY@tok@go\endcsname{\def\PY@tc##1{\textcolor[rgb]{0.53,0.53,0.53}{##1}}}
\expandafter\def\csname PY@tok@gt\endcsname{\def\PY@tc##1{\textcolor[rgb]{0.00,0.27,0.87}{##1}}}
\expandafter\def\csname PY@tok@err\endcsname{\def\PY@bc##1{\setlength{\fboxsep}{0pt}\fcolorbox[rgb]{1.00,0.00,0.00}{1,1,1}{\strut ##1}}}
\expandafter\def\csname PY@tok@kc\endcsname{\let\PY@bf=\textbf\def\PY@tc##1{\textcolor[rgb]{0.00,0.50,0.00}{##1}}}
\expandafter\def\csname PY@tok@kd\endcsname{\let\PY@bf=\textbf\def\PY@tc##1{\textcolor[rgb]{0.00,0.50,0.00}{##1}}}
\expandafter\def\csname PY@tok@kn\endcsname{\let\PY@bf=\textbf\def\PY@tc##1{\textcolor[rgb]{0.00,0.50,0.00}{##1}}}
\expandafter\def\csname PY@tok@kr\endcsname{\let\PY@bf=\textbf\def\PY@tc##1{\textcolor[rgb]{0.00,0.50,0.00}{##1}}}
\expandafter\def\csname PY@tok@bp\endcsname{\def\PY@tc##1{\textcolor[rgb]{0.00,0.50,0.00}{##1}}}
\expandafter\def\csname PY@tok@fm\endcsname{\def\PY@tc##1{\textcolor[rgb]{0.00,0.00,1.00}{##1}}}
\expandafter\def\csname PY@tok@vc\endcsname{\def\PY@tc##1{\textcolor[rgb]{0.10,0.09,0.49}{##1}}}
\expandafter\def\csname PY@tok@vg\endcsname{\def\PY@tc##1{\textcolor[rgb]{0.10,0.09,0.49}{##1}}}
\expandafter\def\csname PY@tok@vi\endcsname{\def\PY@tc##1{\textcolor[rgb]{0.10,0.09,0.49}{##1}}}
\expandafter\def\csname PY@tok@vm\endcsname{\def\PY@tc##1{\textcolor[rgb]{0.10,0.09,0.49}{##1}}}
\expandafter\def\csname PY@tok@sa\endcsname{\def\PY@tc##1{\textcolor[rgb]{0.73,0.13,0.13}{##1}}}
\expandafter\def\csname PY@tok@sb\endcsname{\def\PY@tc##1{\textcolor[rgb]{0.73,0.13,0.13}{##1}}}
\expandafter\def\csname PY@tok@sc\endcsname{\def\PY@tc##1{\textcolor[rgb]{0.73,0.13,0.13}{##1}}}
\expandafter\def\csname PY@tok@dl\endcsname{\def\PY@tc##1{\textcolor[rgb]{0.73,0.13,0.13}{##1}}}
\expandafter\def\csname PY@tok@s2\endcsname{\def\PY@tc##1{\textcolor[rgb]{0.73,0.13,0.13}{##1}}}
\expandafter\def\csname PY@tok@sh\endcsname{\def\PY@tc##1{\textcolor[rgb]{0.73,0.13,0.13}{##1}}}
\expandafter\def\csname PY@tok@s1\endcsname{\def\PY@tc##1{\textcolor[rgb]{0.73,0.13,0.13}{##1}}}
\expandafter\def\csname PY@tok@mb\endcsname{\def\PY@tc##1{\textcolor[rgb]{0.40,0.40,0.40}{##1}}}
\expandafter\def\csname PY@tok@mf\endcsname{\def\PY@tc##1{\textcolor[rgb]{0.40,0.40,0.40}{##1}}}
\expandafter\def\csname PY@tok@mh\endcsname{\def\PY@tc##1{\textcolor[rgb]{0.40,0.40,0.40}{##1}}}
\expandafter\def\csname PY@tok@mi\endcsname{\def\PY@tc##1{\textcolor[rgb]{0.40,0.40,0.40}{##1}}}
\expandafter\def\csname PY@tok@il\endcsname{\def\PY@tc##1{\textcolor[rgb]{0.40,0.40,0.40}{##1}}}
\expandafter\def\csname PY@tok@mo\endcsname{\def\PY@tc##1{\textcolor[rgb]{0.40,0.40,0.40}{##1}}}
\expandafter\def\csname PY@tok@ch\endcsname{\let\PY@it=\textit\def\PY@tc##1{\textcolor[rgb]{0.25,0.50,0.50}{##1}}}
\expandafter\def\csname PY@tok@cm\endcsname{\let\PY@it=\textit\def\PY@tc##1{\textcolor[rgb]{0.25,0.50,0.50}{##1}}}
\expandafter\def\csname PY@tok@cpf\endcsname{\let\PY@it=\textit\def\PY@tc##1{\textcolor[rgb]{0.25,0.50,0.50}{##1}}}
\expandafter\def\csname PY@tok@c1\endcsname{\let\PY@it=\textit\def\PY@tc##1{\textcolor[rgb]{0.25,0.50,0.50}{##1}}}
\expandafter\def\csname PY@tok@cs\endcsname{\let\PY@it=\textit\def\PY@tc##1{\textcolor[rgb]{0.25,0.50,0.50}{##1}}}

\def\PYZus{\char`\_}
\def\PYZob{\char`\{}
\def\PYZcb{\char`\}}

\def\PYZgt{\char`\>}
\def\PYZsh{\char`\#}
\def\PYZpc{\char`\%}

\def\PYZhy{\char`\-}
\def\PYZsq{\char`\'}

\makeatother
 \DeclareUnicodeCharacter{03B1}{\ensuremath{\alpha}}
\DeclareUnicodeCharacter{03C1}{\ensuremath{\rho}}
\usepackage{trimclip}

\DeclareMathOperator*{\argmin}{arg\,min}

\title{Solving MaxSAT with Matrix Multiplication}

\correspondingauthor{dwf@google.com,vinair@google.com}

\reportnumber{} 

\renewcommand{\today}{} 

\author[*,1]{David Warde-Farley}
\author[*,1]{Vinod Nair}
\author{Yujia Li}
\author{Ivan Lobov}
\author{Felix Gimeno}
\author{Simon Osindero}

\affil[*]{Equal contribution}
\affil[1]{Google DeepMind}

\begin{abstract}
We propose an incomplete algorithm for Maximum Satisfiability (MaxSAT) specifically designed to run on neural network accelerators such as GPUs and TPUs. Given a MaxSAT problem instance in conjunctive normal form, our procedure constructs a Restricted Boltzmann Machine (RBM) with an equilibrium distribution wherein the probability of a Boolean assignment is exponential in the number of clauses it satisfies. Block Gibbs sampling is used to stochastically search the space of assignments with parallel Markov chains. Since matrix multiplication is the main computational primitive for block Gibbs sampling in an RBM, our approach leads to an elegantly simple algorithm (40 lines of JAX) well-suited for neural network accelerators. Theoretical results about RBMs guarantee that the required number of visible and hidden units of the RBM scale only linearly with the number of variables and constant-sized clauses in the MaxSAT instance, ensuring that the computational cost of a Gibbs step scales reasonably with the instance size. Search throughput can be increased by batching parallel chains within a single accelerator as well as by distributing them across multiple accelerators. As a further enhancement, a heuristic based on unit propagation running on CPU is periodically applied to the sampled assignments. Our approach, which we term \emph{RbmSAT}, is a new design point in the algorithm-hardware co-design space for MaxSAT. We present timed results on a subset of problem instances from the annual MaxSAT Evaluation's Incomplete Unweighted Track for the years 2018 to 2021. When allotted the same running time and CPU compute budget (but no TPUs), RbmSAT outperforms other participating solvers on problems drawn from three out of the four years' competitions. Given the same running time on a TPU cluster for which RbmSAT is uniquely designed, it outperforms all solvers on problems drawn from all four years. \end{abstract}

\begin{document}

\maketitle
\renewcommand\vec\mathbf
\newcommand\mat\mathbf
\renewcommand\t[1]{{#1}^\mathsf{T}}

\newcommand{\unstructuredunits}{x}

\newcommand{\vis}{v}
\newcommand{\hid}{h}

\newcommand{\UnstructuredUnits}{\vec{\unstructuredunits}}
\newcommand{\Vis}{\vec{\vis}}
\newcommand{\Hid}{\vec{\hid}}

\newcommand{\unstructuredb}{a}
\newcommand{\visb}{d}
\newcommand{\hidb}{b}

\newcommand{\UnstructuredB}{\vec{\unstructuredb}}
\newcommand{\VisB}{\vec{\visb}}
\newcommand{\HidB}{\vec{\hidb}}

\newcommand{\unstructuredw}{s}
\newcommand{\vishid}{w}
\newcommand{\visvis}{z}
\newcommand{\hidhid}{u}

\newcommand{\matrixparam}[1]{\mat{\MakeUppercase{#1}}}
\newcommand{\VisHid}{\matrixparam{\vishid}}
\newcommand{\VisVis}{\matrixparam{\visvis}}
\newcommand{\HidHid}{\matrixparam{\hidhid}}
\newcommand{\UnstructuredW}{\matrixparam{\unstructuredw}}

\newcommand{\VisIndex}{i}
\newcommand{\HidIndex}{j}
\newcommand{\AltVisIndex}{j}
\newcommand{\AltHidIndex}{i}
\newcommand{\ClauseVisIndex}{k}
\newcommand{\ClauseHidIndex}{m}
\newcommand{\ClauseIndex}{c}
\newcommand{\IndexTableRowIndex}{\ClauseIndex}
\newcommand{\IndexTableColIndex}{\ClauseVisIndex}

\newcommand{\BatchIndex}{b}
\newcommand{\BatchSize}{B}
\newcommand{\NumVis}{N}
\newcommand{\NumHid}{M}
\newcommand{\NumClauseVis}{K}
\newcommand{\NumClauseHid}{L}
\newcommand{\NumClauses}{C}

\newcommand{\Energy}{E}
\newcommand{\FreeEnergy}{F}
\newcommand{\Prob}{P}
\newcommand{\PartitionFunc}{\mathcal{Z}}

\newcommand{\Params}{\mathbf{\theta}}

\newcommand{\EnergyP}{\Energy_\Params}
\newcommand{\FreeEnergyP}{\FreeEnergy_\Params}
\newcommand{\ProbP}{\Prob_\Params}
\newcommand{\PartitionFuncP}{\PartitionFunc_\Params}
\newcommand{\VisHidStack}[1]{\VisHid^{({#1})}}

\newcommand{\ClauseVis}[1]{\Vis^{({#1})}}
\newcommand{\ClauseHid}[1]{\Hid^{({#1})}}
\newcommand{\ClauseHidB}[1]{\HidB^{({#1})}}
\newcommand{\ClauseVisB}[1]{\VisB^{({#1})}}
\newcommand{\ClauseHidCols}[1]{\vec{u}_{#1}}
\newcommand{\SumClauses}{\sum_{\ClauseIndex=1}^\NumClauses}
\newcommand{\ClauseParams}[1]{{\Params^{({#1})}}}

\newcommand{\Expectation}{\mathbb{E}}
\newcommand{\Naturals}{\mathbb{N}}
\newcommand{\Integers}{\mathbb{Z}}
\newcommand{\Reals}{\mathbb{R}}

\newcommand{\HidIndexVec}{\vec{r}}
\newcommand{\OneBasedIndexTable}{\mat{T}}
\newcommand{\OneBasedIndexTableRow}{\mathbf{t}}
\newcommand{\SignAndIndexTransposedTable}{\mathcal{T}}
\newcommand{\OneBasedIndexTableElem}{t}
\newcommand{\OneHotExpansionElem}{q}
\newcommand{\OneHotExpansion}{\mat{Q}}
\newcommand{\OneHotExpansionSlice}[1]{\mat{Q}_{#1}}

\newcommand{\SubVector}[3]{\t{({#1}_{#2})}_{{#2} \in {#3}}}

\newcommand{\LogitContribs}[1]{\ell^{({#1})}}
\newcommand{\LogitContrib}[2]{\ell_{#1}^{({#2})}}

\newcommand{\PartialWrtParams}[1]{
\frac{
\partial {#1}
}{
\partial \Params
}
}

\newcommand{\BooleanAnd}{AND}
\newcommand{\BooleanOr}{OR}
\newcommand{\OrGate}{\BooleanOr-gate}
\newcommand{\ComplexityClass}[1]{\textbf{#1}}
\newcommand{\ClassP}{\ComplexityClass{P}}
\newcommand{\NP}{\ComplexityClass{NP}}
\newcommand{\AXP}{\ComplexityClass{AXP}}
\newcommand{\BooleanFunctionInputElem}{x}
\newcommand{\BooleanFunctionInput}{\vec{\BooleanFunctionInputElem}}
\newcommand{\BooleanFunctionInputWeight}{\mathcal{\MakeUppercase{\BooleanFunctionInputElem}}}
\newcommand{\PolarityDiag}{\mat{\Lambda}}
\newcommand{\PolarityDiagElem}{\lambda}

\newcommand{\Queue}{\mathcal{Q}} 
\section{Introduction}
\label{sec:intro}
Given a Boolean formula $f(\vec{x})$ over a set of $N$ variables $\vec{x}\in\{\text{True},\text{False}\}^N$, \emph{Boolean Satisfiability} (SAT) is the problem of deciding if there exists an assignment of $\vec{x}$ such that $f(\vec{x}) = \text{True}$. It was the first problem shown to be \NP-complete~\citep{Cook:1971}, and SAT solvers have a wide range of practical applications~\cite{marques-silva2008SATapplications, biere2009sat-handbook}. Typically $f(\vec{x})$ is in \emph{Conjunctive Normal Form} (CNF), which is a conjunction (\BooleanAnd) of \emph{clauses}, where each clause is a disjunction (\BooleanOr) of a subset of variables or their negations. \emph{Maximum Satisfiability} (MaxSAT) \cite{bacchus2021MaxSAT} generalizes SAT, requiring determination of the maximum number of clauses of $f(\vec{x})$ that can be satisfied with a single assignment. A weighted variant assigns weights to clauses, with solvers aiming to maximize the total weight of satisfied clauses. An \emph{incomplete} algorithm~\cite{kautz2021incomplete} for MaxSAT aims to find a good solution quickly without attempting to prove its optimality.

State-of-the-art solvers for SAT and MaxSAT have been designed with decades of research to run efficiently on CPUs (see, e.g., annual SAT Competitions \cite{sat-comp} since 2002 and MaxSAT Evaluations~\cite{maxsat-eval} since 2006, both of which evaluate CPU-based solvers). Both algorithmic and hardware improvements have enabled solvers to scale to larger problems~\cite{kotthoff2018quantifying, fichte2020timeleapSAT} and solve them faster, with one study estimating that both kinds of improvements have contributed roughly equally~\cite{fichte2020timeleapSAT}. However, as Moore's law slows down, CPU improvements can no longer be relied upon to deliver significant performance gains for solvers.

Special-purpose hardware accelerators are a potential path forward. In particular, neural network accelerators such as Tensor Processing Units (TPUs) \cite{jouppi2017tpus} and Graphics Processing Units (GPUs), provide immense throughput for specific operations such as matrix multiplication. For example, the TPUv4i chip has a peak performance of 138 TeraFLOPS, with a cluster of 1024 chips providing a peak performance of $> 100$ PetaFLOPS~\cite{jouppi2021tenlessons}. Both GPUs and TPUs have also shown 5-10$\times$ performance improvement over the past several years~\cite{nvidia2022h100, jouppi2021tenlessons}, offering a tempting alternative to the diminishing returns of Moore's law. How to best implement a MaxSAT solver on neural network accelerators is an open question. Given the CPU-centric design of state-of-the-art solvers and their use of search algorithms that are not easy to express as matrix multiplication (e.g., tree search), attempting a direct retargeting of such solvers is unlikely to be effective~\cite{SOHANGHPURWALA2017hwsat}.

\begin{figure*}
    \centering
    \includegraphics[trim={0.25cm, 3.5cm, 0cm, 4cm}, clip, width=\linewidth]{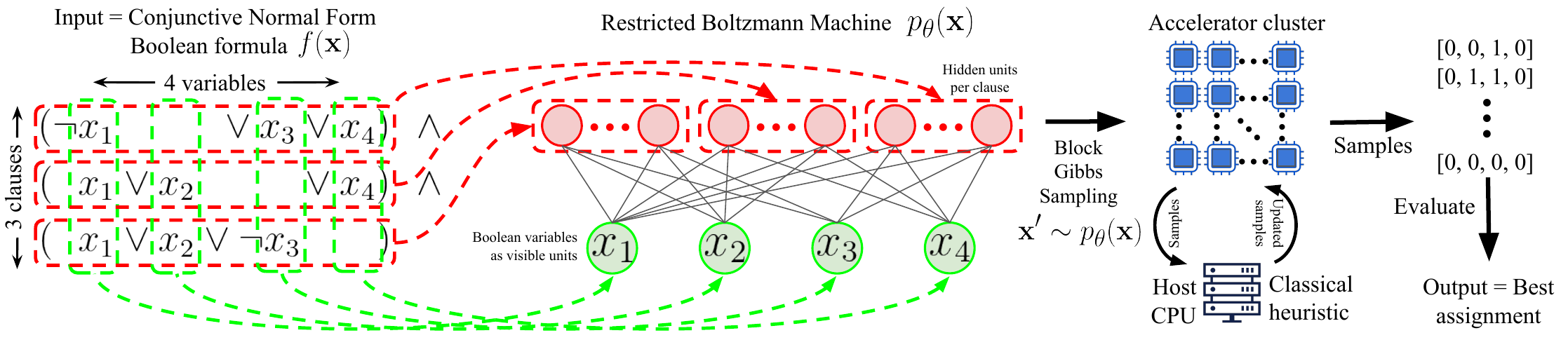}
\caption{\small Given a Boolean formula $f(\vec{x})$ in conjunctive normal form with $N$ variables and $C$ clauses ($N=4$, $C=3$ above), RbmSAT constructs a Restricted Boltzmann Machine (RBM) $p_{\theta}(\vec{x})$ with $\vec{x}$ as its binary visible units, weights $\theta$, and $O(K)$ binary hidden units per clause of size $K$ ($K=3$ above). By construction $p_{\theta}(\vec{x})$ gives higher probability to an assignment of $\vec{x}$ that satisfies more clauses. RBM's block Gibbs sampling is used to stochastically search for high probability assignments, which requires matrix multiplication as the main operation and can be done scalably on a TPU/GPU cluster. Samples are periodically updated with a heuristic (running on a CPU) that unassigns values for a subset of variables in a Gibbs chain and re-assigns them values logically implied by the assigned variables. The sampling iterations continue till the time limit, when the best assignment seen so far is output.}
    \label{fig:rbmsat_intro}
\end{figure*}

We propose an incomplete algorithm for unweighted MaxSAT uniquely designed for neural network accelerators (Figure~\ref{fig:rbmsat_intro}). The key idea is elegantly simple: given $f(\vec{x})$ in CNF with $N$ variables and $C$ clauses, a distribution $p_{\theta}(\vec{x})$ parameterized by $\theta$ is constructed as a \emph{Restricted Boltzmann Machine} (RBM)~\cite{smolensky1986information,freund1991unsupervised, hinton2002training}, an energy-based probabilistic model for high-dimensional distributions~\citep{leroux2008representational,montufar2011refinements}. $\theta$ is chosen such that the more clauses an assignment satisfies, the higher its probability under $p_{\theta}(\vec{x})$. Theoretical results on the representational capabilities of RBMs~\cite{martens2013rbms, gu2019rbms} guarantee that such an RBM can be constructed with $N$ visible units and $O(KC)$ hidden units for clause size $K$ (section \ref{sec:approach}). Gibbs sampling for $p_{\theta}(\vec{x})$ is used to find highly probable, and thus highly satisfying, variable assignments. The RBM's bipartite graph structure between visible and hidden units enable efficient block Gibbs sampling using matrix multiplication (section \ref{subsec:bm-background}). The result is a scalable stochastic search algorithm for MaxSAT well-suited for a TPU/GPU cluster. We further augment the search by periodically improving the sampled assignments with a simple classical heuristic (section \ref{subsec:unit_prop}). We refer to the algorithm as \emph{RbmSAT}. See  section~\ref{subsec:algorithm}, figure~\ref{fig:code} for a simple 40-line JAX implementation.

This paper introduces the concepts behind RbmSAT and presents an initial evaluation on a subset of unweighted problem instances (selected by size to fit our implementation constraints, see section \ref{sec:evaluation}) from the Incomplete Unweighted Track of four recent annual MaxSAT Evaluations (2018-2021, see \cite{maxsat-eval}). Following the methodology of the MaxSAT Evaluations, we compare RbmSAT to the participating solvers by the \emph{average incomplete score} (section \ref{sec:evaluation}) at a wall clock time limit of 300 seconds. The results (section \ref{subsec:results}) show that in a controlled comparison using the same \emph{CPU}, memory, and time budget as the other solvers, RbmSAT achieves the best average incomplete score for problems drawn from three out of the four years. In an uncontrolled comparison for the same 300 second time limit but that provides RbmSAT the advantage of a 64-TPU cluster, it outperforms all other solvers on problems drawn from all four years. Since both comparisons are on a subset of instances that omit the largest ones, and the second comparison is across solvers that run on very different hardware (CPUs and TPUs) that are inherently difficult to compare, we do \emph{not} claim that RbmSAT is a state-of-the-art solver. We nonetheless believe these results provide evidence for the viability of RbmSAT that justifies further research.

Our contributions are the following:
\begin{enumerate}
    \item We propose RbmSAT, a MaxSAT algorithm specifically designed for neural network accelerators. 
    \item We describe a practical implementation of RbmSAT that easily scales parallel Gibbs sampling with more accelerators.
    \item On the Incomplete Unweighted Track of MaxSAT Evaluations 2018 to 2021, RbmSAT outperforms all participating solvers on a subset of problem instances using a 64-TPU cluster in a comparison that controls for wall clock time but not compute, and remains competitive even when deployed on a traditional CPU.
\end{enumerate}

 \section{Related Work}
\label{sec:related_work}
\vspace{-0.1in}
The literature on SAT and MaxSAT is vast~\cite{biere2021handbook}. Many modern MaxSAT solvers rely on SAT solvers internally by iteratively augmenting the input problem with extra constraints and applying a SAT solver to improve upon the previous solution~\cite{bacchus2021MaxSAT}. Popular approaches for SAT broadly belong to two classes: 1) tree search based on the Conflict Driven Clause Learning (CDCL) algorithm (see, e.g. \cite{MarquesSilva2009ConflictDrivenCL}), and 2) stochastic local search (see, e.g., \cite{hoos2000LSforSAT}). It is not immediately obvious how such solvers can make efficient use of neural network accelerators~\cite{SOHANGHPURWALA2017hwsat}. Approaches that use large-scale parallel CPU compute, such as Cube and Conquer~\cite{heule2012cube_and_conquer} and entries in the Parallel Track of SAT competitions, typically retrofit a conventional approach with parallel compute, e.g. by processing sub-trees of a search tree in parallel, but they have not been broadly adopted. 
Hardware acceleration (e.g., FPGA) has been used to speed up \emph{unit propagation} which is a computational bottleneck in CDCL-based solvers~\cite{SOHANGHPURWALA2017hwsat}. The classical heuristic upon which RbmSAT relies (section~\ref{subsec:unit_prop}) is based on unit propagation and can also benefit from such accelerators.

Ising models have been used to encode combinatorial optimization problems such that low energy configurations of the Ising model correspond to high quality solutions, see, e.g., \cite{lucas2014IsingforNP}. Early examples are Hopfield networks for the Travelling Salesman Problem~\citep{hopfieldtank1985tsp} and (general) Boltzmann Machines for SAT and MaxSAT \cite{anjou1993bmsat}. Approaches for simulating Ising models using GPUs~\cite{cook2019isinggpu} and TPUs~\citep{yang2019isingtpu} have been proposed. RBM's bipartite graph structure differs from the lattice structure of a 2D Ising model, and admits efficient block Gibbs sampling using matrix multiplication, making it a natural fit for neural network accelerators unlike Ising models and Boltzmann Machines (without connectivity constraints). It is also possible to encode an Ising model's energy function as an RBM's free energy function using the results from~\cite{martens2013rbms, gu2019rbms} and extend our approach to Ising models.

Survey Propagation~\citep{braunstein2005surveyprop} constructs a factor graph for a Boolean Satisfiability problem with variables as one set of nodes and clauses as the other.
The defined distribution assigns zero probability to unsatisfying assignments and uniform probability to satisfying assignments.
Message passing is used to estimate the marginal distributions of the variables.
Those variables with high marginal probability of being 0 or 1 are fixed to their maximum \textit{a posteriori} values, and the resulting (typically much) smaller sub-problem is solved using other approaches. The distribution defined by RbmSAT assigns non-zero probability to all assignments, but with the probability of an assignment exponential in the number of satisfied clauses. It matches the factor graph distribution only at the zero temperature limit.

Several works learn search policies for SAT that generalize across problem instances (e.g. \cite{selsam2019NeuroSAT, selsam2019guiding, amizadeh2019LearningTS, yolcu2019learnLS, li2018guidedTreeSearch}).
In this work we do not learn a search policy, but such methods are complementary and can be combined with RbmSAT (e.g.\  by learning to generate good initial assignments for RbmSAT).

 \section{Background}

\subsection{Restricted Boltzmann Machines}
\label{subsec:bm-background}

\emph{Boltzmann Machines}~\citep{hinton1986learning} are undirected graphical models defined by the energy function $E_{\theta}(\vec{x}) = -\t{\UnstructuredUnits}\UnstructuredW\UnstructuredUnits-\t{\UnstructuredB}\UnstructuredUnits$, on variables $\vec{x} \in \{0,1\}^D$, parameterized by $\theta = \{\vec{S},\vec{a}\}$ where $\vec{S} \in \mathbb{R}^{D \times D}$ is symmetric and $\vec{a} \in \mathbb{R}^D$. It defines a probability distribution $p_{\theta}(\vec{x}) = \exp(-E_{\theta}(\vec{x}))/Z(\theta)$, where $Z(\theta) = \sum_{\vec{x}'}\exp(-E_{\theta}(\vec{x}'))$ is the partition function. For large $D$ computing $Z(\theta)$ requires an intractable summation, making $p_{\theta}(\vec{x})$ and its gradient (needed to learn $\theta$) intractable as well.

\emph{Restricted Boltzmann Machines} (RBMs)~\citep{smolensky1986information} are a special case in which $\vec{x}$ consists of observed and latent variables and the model imposes a bipartite connectivity structure on the two groups: no connections are permitted between two observed variable elements of $\vec{x}$ \emph{or} between two latent variable elements of $\vec{x}$. Denoting the observed variables as \emph{visible units} $\vec{v} \in \{0,1\}^{N}$ and the latent variables as \emph{hidden units} $\vec{h} \in \{0,1\}^{N_h}$, the energy function can now be written as $E_{\theta}(\vec{v},\vec{h}) = -\t{\Vis}\VisHid\Hid -\t{\Vis}\VisB - \t{\Hid}\HidB$, where $\theta = \{\vec{W},\vec{d},\vec{b}\}$, $\vec{W} \in \mathbb{R}^{N \times N_h}, \vec{d} \in \mathbb{R}^{N}$ and $\vec{b} \in \mathbb{R}^{N_h}$. The \emph{free energy} is the unnormalized negative log probability of $\vec{v}$, and is defined as $\FreeEnergyP(\Vis) = -\log\sum_{\Hid'}\exp\left(-\EnergyP(\Vis,\Hid')\right)$. Given the RBM's bipartite graph structure this simplifies to the tractable expression
\begin{align}
\label{eq:free-energy}
\FreeEnergyP(\Vis) = -\t{\VisB}\Vis - \sum_{j=1}^{N_h} \log \left[1 + \exp\left(\hidb_j + \sum_{k=1}^N \vishid_{kj}\vis_k \right)\right]
\end{align}
The RBM's structure also renders the joint conditional distributions $p_{\theta}(\vec{v}|\vec{h})$ and $p_{\theta}(\vec{h}|\vec{v})$ tractable. This enables parallel \emph{block} Gibbs sampling. The transition from a sample $\tilde{\vec{v}}_t$ at step $t$ to the sample $\tilde{\vec{v}}_{t+1}$ at $t+1$ is given by
\begin{align}
\label{eq:hidden-conditional}
\tilde{\vec{h}}_t & \sim p_{\theta}(\Hid\ |\ \Vis = \tilde{\vec{v}_t})  = \sigma(\HidB + \t{\VisHid}\tilde{\Vis_t})\\
\label{eq:vis-conditional}
\tilde{\vec{v}}_{t+1} & \sim  p_{\theta}(\Vis\ |\ \Hid = \tilde{\Hid_t}) = \sigma(\VisB + \VisHid\tilde{\Hid_t})
\end{align}
where $\sigma(x) = (1 + e^{-x})^{-1}$ is the logistic function, applied element-wise.
RBM sampling is well suited to neural network accelerators as the majority of operations necessary to compute~\eqref{eq:hidden-conditional} and~\eqref{eq:vis-conditional} are expressible as matrix multiplications.

RBMs can be viewed as \emph{products of experts}~\citep{hinton2002training}: each hidden unit (and visible unit bias $\visb_\HidIndex$) contributes a multiplicative constraint on the unnormalized marginal distribution of visible units. The product of several RBMs is itself, therefore, an RBM. We use this property to construct an RBM representing a Boolean formula in CNF as a product of RBMs, each representing one clause in the formula. \section{RbmSAT}
\label{sec:approach}

\subsection{MaxSAT as Minimizing RBM Free Energy}
Theoretical results on the representational efficiency of RBMs~\cite{martens2013rbms, gu2019rbms} form the basis of our approach. In particular, \cite{martens2013rbms} proves that any distribution $p(\vec{v})$ over $\vec{v}\in\{0,1\}^N$ with a \emph{symmetric} free energy function can be represented by an RBM $p_{\theta}(\vec{v})$ with $N$ visible units and $O(N^2)$ hidden units. A function $g(\vec{v})$ is symmetric if and only if $g(\vec{v}) = q(\mathcal{V})$ where $\mathcal{V} = \sum_{i=1}^N{v_i}$, i.e., $g(\cdot)$ depends only on the number of ones in its input. Subsequently, \cite{gu2019rbms} improves the result to $O(N)$ hidden units. Both papers provide analytic procedures for constructing such an RBM and its parameters given the desired free energy function. 

\begin{figure*}
    \centering
    \includegraphics[trim={0.6cm, 0.7cm, 0.2cm, 0.7cm},width=\linewidth]{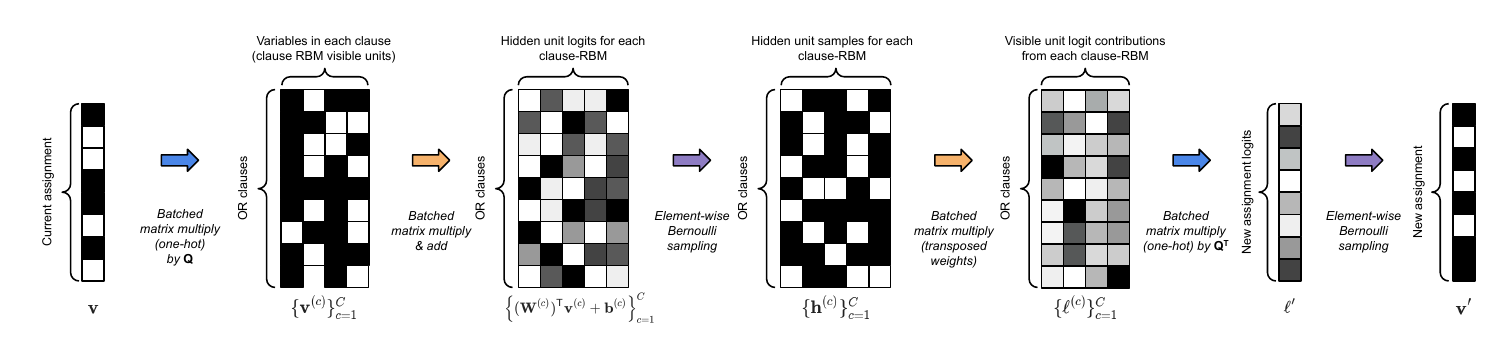}
\caption{\small Accelerator-side computation of a single Gibbs sampling chain broadly consists of gather/scatter operations (blue arrows), clause-RBM batched weight matrix multiplications (orange arrows) and element-wise Bernoulli sampling (purple arrows). An initial assignment vector is gathered into a matrix of clause-RBM inputs via batched matrix multiplication, while a second batched matrix multiply, along with the addition of a per-clause bias vector $\mathbf{b}^{(c)}$ yields logits for the conditional distribution over clause-RBM hidden units. A configuration of the hidden units is sampled and the transpose of the two aforementioned matrix multiplies are applied yielding a set of visible unit logits, from which a new assignment of the variables is sampled. In practice, multiple chains are run in parallel on a mini-batch of variable assignments.}
    \label{fig:rbmsat_overview}
\end{figure*}

\paragraph{OR gate as an RBM:} Boolean OR is a symmetric function. To see this let 0 denote $\mathrm{False}$ and 1 denote $\mathrm{True}$, and note that for an input $\vec{v}\in\{0,1\}^N$, the gate's output depends only on whether $\sum_{i=1}^N{v_i} > 0$. Let $p_{\mathrm{OR}}(\vec{v})$ be a distribution with a free energy function $F_{\mathrm{OR}}(\vec{v})$ such that $F_{\mathrm{OR}}(\vec{v}=\vec{0}) = 0$, corresponding to the only unsatisfying input, and $F_{\mathrm{OR}}(\vec{v} \neq \vec{0}) = F_s$ where $F_s < 0$ is the free energy for all other inputs. $p_{\mathrm{OR}}$ thus assigns equal probability to all satisfying assignments and a lower probability to the single unsatisfying assignment. The corresponding $p_{\mathrm{OR}}$ We use the results of~\cite{martens2013rbms, gu2019rbms} to construct an RBM $p_{\theta}(\vec{v}) = p_{\mathrm{OR}}(\vec{v})$. $F_s$ acts as a temperature parameter: $F_s=0$ yields a uniform distribution (the unsatisfying assignment has equal probability to any satisfying assignment), and lower values reduce $p_{\theta}(\vec{v}=\vec{0})$.

Given the parameters $\VisHid, \HidB$ (visible biases $\VisB$ are not required) of an RBM for a canonical OR gate that does not negate any inputs, the parameters of an RBM for an OR gate involving negations of one or more inputs can be obtained by a simple transformation
\begin{align}
\label{eq:negation-transform}
\begin{array}{c c}
\VisHid' = \PolarityDiag \VisHid; & 
\HidB' = \HidB + \textstyle{\frac12}\left(1 - \mathrm{diag}\left(\PolarityDiag\right)\right)\VisHid
\end{array}
\end{align}
where $\PolarityDiag$ is a diagonal matrix with its $k^{th}$ diagonal element $\PolarityDiagElem_{\ClauseVisIndex\ClauseVisIndex} \in \{-1, 1\}$ set to $-1$ if the $k^{th}$ input to the OR gate is negated.

\paragraph{CNF Boolean formula as an RBM:} A ``formula-RBM'' can be constructed for a Boolean formula in CNF by a) constructing a ``clause-RBM'' for each of the $C$ clauses (an OR gate) with $L$ hidden units and the $K$ variables in the clause as its visible units, and b) taking the product of the clause-RBMs by concatenating their hidden units (for a total of $C \cdot L$ hidden units) and merging the overlapping sets of variables participating in each clause into a vector of $N$ visible units. The free energy of the product is simply the sum of the free energies of the clause-RBMs. Each satisfied clause contributes $F_s < 0$ to the total free energy, with any unsatisfied clauses contributing zero. The total free energy is thus monotonically decreasing, and probability monotonically increasing, in the number of satisfied clauses of an assignment.

Let $\theta_f = \{\vec{W}_f, \vec{d}_f, \vec{b}_f\}$ be the parameters of the formula-RBM. Let $\OneBasedIndexTable \in (\Naturals \cup \{0\})^{\NumClauses \times \NumClauseVis}$. We have $\OneBasedIndexTableElem_{\IndexTableRowIndex\IndexTableColIndex} \leq  \NumVis$ $\forall c,k$, and if $t_{\IndexTableRowIndex\IndexTableColIndex} = 0$ this indicates that group $\IndexTableRowIndex$ contains fewer than $\NumClauseVis$ variables (i.e. the $0$ values are used as padding).
Let $\VisHidStack{\ClauseIndex}_f$ be the (in general, non-contiguous) sub-matrix of $\VisHid_f$ with rows corresponding to the non-zero elements of $\OneBasedIndexTable_{\ClauseIndex,\cdot}$, and columns $\ClauseHidCols{\ClauseIndex}~=~\t{((\ClauseIndex~-~1)\NumClauseHid, (\ClauseIndex~-~1)\NumClauseHid~+~1, \ldots, \ClauseIndex\NumClauseHid)}$.
Similarly, define (for $j \neq 0$)
\begin{align*}
\begin{array}{rcl rcl}
\ClauseVis{\ClauseIndex} & = & \SubVector{\vis}{j}{\OneBasedIndexTable_{\ClauseIndex,\cdot}} &
\ClauseVisB{\ClauseIndex} & = & \SubVector{\visb}{j}{\OneBasedIndexTable_{\ClauseIndex,\cdot}} \\
\ClauseHidB{\ClauseIndex} & = &  \SubVector{\hidb}{j}{\ClauseHidCols{\ClauseIndex}} &
\ClauseHid{\ClauseIndex} & = & \SubVector{\hid}{j}{\ClauseHidCols{\ClauseIndex}}
\end{array}
\end{align*}
Denoting $\theta_f^{(c)} = \{\VisHidStack{\ClauseIndex}_f,\ClauseVisB{\ClauseIndex}_f,\ClauseHidB{\ClauseIndex}_f\}$, the free energy of the formula-RBM is given by:
\begin{align}
F_{\theta_f}(\Vis)
= & \SumClauses
F_{\theta^{(c)}_f}(\ClauseVis{\ClauseIndex})
\end{align}

\paragraph{Gibbs sampling in a formula-RBM:} Given the groups $\ClauseVis{1}, \ClauseVis{2}, \ldots, \ClauseVis{\NumClauses}$, sampling values for $\Hid$ amounts to sampling each disjoint group of units $\ClauseHid{1}, \ClauseHid{2}, \ldots, \ClauseHid{\NumClauses}$ from its respective clause-RBM:
\begin{align}
\label{eq:sparse_h_given_v}
p_{\theta_f}(\Hid\ |\ \Vis) = & \prod_{\ClauseIndex=1}^\NumClauses p_{\theta^{(c)}_f}\left(\ClauseHid{\ClauseIndex} \ | \ \ClauseVis{\ClauseIndex}\right)
\end{align}
Sampling $\Vis$ given $\Hid$ poses a greater challenge, as every visible unit $\vis_\VisIndex$ may participate in many clause-RBMs. Let 
$\LogitContribs{\ClauseIndex} = \vec{W}_f^{(c)} \ClauseHid{\ClauseIndex}$. The logits of the conditional distribution decompose as a sum of contributions from each $\LogitContribs{\ClauseIndex}$ that contains $\vis_\VisIndex$, added to a global bias $\visb_{\VisIndex}$:
\begin{align}
\label{eq:sparse_v_given_h}
p_{\theta}(\vis_\VisIndex \ | \ \Hid) = \sigma\left(
   \visb_\VisIndex + 
   \SumClauses
   \sum_{\ClauseVisIndex=1}^\NumClauseVis
   \delta_{\VisIndex,\OneBasedIndexTableElem_{\ClauseIndex\ClauseVisIndex}}
   \LogitContrib{\ClauseVisIndex}{\ClauseIndex}
\right)
\end{align}
with $\delta_{i,j} = 1$ for $i = j$ and 0 for $i \neq j$. A practical implementation of RBM operations with these \emph{a priori} connectivity constraints need never instantiate the $N \times CL$ weight matrix $\VisHid_f$, which will be very large for large $N$ and $C$, and sparse for $\NumClauseVis \ll \NumVis$. Instead, each sub-matrix $\vec{W}_f^{(c)}$ can be stored contiguously and zero entries of $\vec{W}_f$ represented only implicitly.

    \paragraph{Learning OR gate RBM parameters:} The RBM parameters for an OR gate can be computed analytically using the procedure in \cite{martens2013rbms} or \cite{gu2019rbms}. However, we have observed empirically that learning them by directly fitting the free energy function for an OR gate yields RBMs for which Gibbs sampling mixes faster, which is crucial for efficient search. The analytically computed weights have high magnitude, which slows mixing. Typically SAT and MaxSAT problem instances have a small enough number of variables $K$ in a single clause (e.g., we assume $K <= 7$ in this work) that all possible $2^K$ clause inputs can be exhaustively enumerated, in which case the RBM parameters $\theta$ can be learned by regressing the free energy to the target values as $\theta_{\FreeEnergy_s} = \argmin_{\Params} \| \FreeEnergyP(\vec{0})\|^2 + \sum_{\Vis \neq \vec{0}} \|\FreeEnergyP(\Vis) - \FreeEnergy_s\|^2$. 
    
    An equivalent concatenative approach to constructing RBMs for representing Boolean circuits from smaller pre-trained RBMs was proposed by~\cite{patel2019combining}, who perform learning of the constituent RBMs using the contrastive divergence approximation to the maximum likelihood gradient.
    This unfortunately introduces unnecessary noise and does not optimize the RBM log likelihood (or, indeed, any objective function~\citep{sutskever2010convergence}).
    If $K$ is too large to allow for enumeration, it is possible to relax the fully deterministic learning described above while still avoiding MCMC gradient approximations by sampling satisfying assignments uniformly at random on each learning step, but we have not yet empirically examined this possibility.

\subsection{Stochastic Search with Gibbs Sampling}
Performing stochastic search consists of instantiating parallel Markov chains, initialized from a uniform distribution over assignments, and performing block Gibbs sampling as explained above. We use an ensemble of values for the temperature $F_s$, train an OR gate RBM for each temperature, and instantiate and simulate all corresponding formula-RBMs in parallel. The clause satisfaction count of an assignment can be computed at each step with very little overhead, as explained below. We retain the best assignment seen by any Markov chain, and output the best identified assignment once we have exhausted the time limit.

\subsection{Scalable TPU Implementation}
\label{subsec:implementation}
We implement our approach in JAX~\cite{jax2018github} targeting TPUs~\cite{cloud-tpu-description}, proprietary ASICs developed by Google to accelerate neural network workloads, which rely heavily on matrix multiplication. JAX programs are just-in-time compiled using the XLA (Accelerated Linear Algebra)~\cite{xla2017} domain-specific compiler. A simplified version of the RbmSAT code is provided in Figure~\ref{fig:code} in the Appendix. 

We perform block Gibbs sampling in multiple RBMs trained with different OR gate satisfying free energy targets synchronously on slices of 64 TPUv4 chips containing 128 TPU cores. 
At a high level, performing a single full round of Gibbs sampling involves \textbf{gathering} variables for each clause, \textbf{sampling hidden states} associated with each clause, \textbf{computing logit contributions} for each participating variable of each clause, \textbf{summing} the logit contributions across clauses with a scatter-add operation, and \textbf{sampling new visible states}.

Weights of the RBMs are assembled from pretrained weights for a canonical OR gate with a suitable maximum clause size $\NumClauseVis$ and the desired free energy target, and transformed as needed according to~\eqref{eq:negation-transform}. A clause with fewer than $\NumClauseVis$ participating variables can be padded to length $\NumClauseVis$ by repeating one of its variables with the same polarity, which leaves the set of satisfying assignments invariant.

For the gather and scatter operations, our implementation stores a table $\SignAndIndexTransposedTable \in \Integers^{\NumClauseVis \times \NumClauses}$ of signed integers\footnote{Typically $\NumClauseVis << \NumClauses$, storing this table variable-major, clause-minor incurs less losses to padding given the TPU's tiled memory model, which is allocated in blocks of $8 \times 128$.}, with $\OneBasedIndexTable = \t{(\mathrm{sign}(\SignAndIndexTransposedTable)\odot\SignAndIndexTransposedTable)}$, and signs representing the polarity (negation) of each position in the clause, i.e.\ a column $\t{(1, 2, -3, 4)}$ represents the clause $v_1 \vee v_2 \vee \neg v_3 \vee v_4$. Both the gather and scatter-add operations can be implemented via batched matrix multiplication (via XLA's \texttt{DotGeneral} primitive, exposed by JAX's \texttt{einsum} operator) on the TPU's matrix multiplication unit (MXU) by transforming $\SignAndIndexTransposedTable$ into a 3-dimensional one-hot representation $\OneHotExpansion \in \{0, 1\}^{\NumClauses\times\NumVis\times\NumClauseVis}$ where $\OneHotExpansionElem_{\ClauseIndex\VisIndex\ClauseVisIndex} = \delta_{\VisIndex,\OneBasedIndexTableElem_{\ClauseIndex\ClauseVisIndex}}$.
Conveniently, the XLA compiler fuses the creation of this intermediate array (implemented as a broadcasted integer equality test) with the \texttt{DotGeneral} operation, eliminating the memory cost of instantiating $\OneHotExpansion$ explicitly.
To perform a gather, a batch of assignments of size $\BatchSize \times \NumVis$ is right-multiplied by each slice $\mat{Q}_{\ClauseIndex,\cdot,\cdot}$, yielding a result of size $\BatchSize \times \NumClauses \times \NumClauseVis$. A scatter-add is performed, conceptually, by right-multiplying each set of logit contributions $\LogitContribs{\ClauseIndex}$, for each batch item and each clause $\ClauseIndex$, by $\t{\mat{Q}}_{\ClauseIndex,\cdot,\cdot}$. Storing the polarities in $\SignAndIndexTransposedTable$ is not necessary for sampling, but storing them on device in this manner allows for efficient evaluation of the number of satisfying clauses by performing a parallel reduction following the gather step by comparing thresholded signs of $\SignAndIndexTransposedTable$ to the Boolean values in each gathered clause, performing the OR operation for each clause, treating the Boolean result as integers in $\{0, 1\}$, and summing the result for each chain state.

\subsection{Unit Propagation-based Heuristic}
\label{subsec:unit_prop}
We augment the Gibbs sampling-based stochastic search with a heuristic proposed in \cite{si2019prioritized}. It is based on \emph{unit propagation}, which attempts to extend a partial assignment of the variables for a formula by computing the partial assignment's logical implications for the unassigned variables. This is a sequential procedure: an initial partial assignment may get extended by implied values for some unassigned variables, which may further imply values for more variables, and so on. For example, in the expression $(a \vee b \vee c) \wedge (\neg c\vee \neg d)$, a partial assignment $a = b = \mathrm{False}$ implies $c$ must be $\mathrm{True}$ for the first clause to be true. However, $c$ is negated in the second clause, which implies a value of $\mathrm{False}$ for $d$.

We rank each variable $\vis_\VisIndex$ by a moving average of the conditional variance $\nu_\VisIndex = \rho_\VisIndex(1 - \rho_\VisIndex)$ of the Bernoulli distribution with parameter $\rho_\VisIndex = p_{\vec{\theta}_f}(v_i = 1\ |\ \Hid)$. The variables with the lowest variance are selected to be unassigned and recomputed using unit propagation. The intuition is that variables with the lowest average conditional variance for a given Markov chain flip state infrequently, which could indicate that the chain is ``stuck'' due to such variables. Starting from a partial assignment that excludes such variables and applying unit propagation can result in a change in value for these ``stuck'' variables, and allow the assignment to make highly non-local moves. The $\lfloor\frac{1}{2}\NumVis\rfloor$ variables with the \emph{highest} value of $\nu_\VisIndex$ are held fixed and the rest are updated. See \cite{si2019prioritized} for more details. Since unit propagation does not use matrix multiplication, it is not well-suited to run on a TPU. The heuristic is applied periodically, with the current samples sent to the host for CPU computation, and the updated samples returned to resume sampling. Results show that the heuristic significantly improves the performance of RbmSAT.

 \section{Evaluation}
\label{sec:evaluation}

\paragraph{Datasets:} We use publicly available instances from the annual MaxSAT Evaluations for the years 2018 to 2021. Since RbmSAT is an incomplete solver and currently does not handle weighted clauses, we use instances from the Incomplete Unweighted Track of each Evaluation\footnote{\hyperlink{https://maxsat-evaluations.github.io/2021/tracks.html}{https://maxsat-evaluations.github.io/2021/tracks.html}}. As a first step towards demonstrating the usefulness of our approach, we select a subset of instances with maximum clause size $\leq7$, number of variables $\leq10000$, and number of clauses $\leq100000$. We chose this subset such that the parameters of the formula-RBM and a batch of assignments would fit in the memory of a single TPU core for all assignments, in order to ease experimentation and resource allocation decisions. Larger instances can be tackled by sharding clauses across multiple accelerators at the cost of a cross-device reduction at every sampling step, but optimizing per-instance Gibbs sampling throughput becomes more challenging in this regime. With regard to clause size in particular, instances with large maximum clause size tend to have few large and many small clauses, and thus a fully homogeneous formula-RBM implementation of the form described here would incur significant memory overhead due to padding smaller clauses to the size of the largest clause through variable repetition. This could be addressed by clustering clauses by size and essentially running several independent formula-RBMs which are coupled by summing their visible logits, at the cost of higher implementational complexity.

After selecting problem instances according to the above size criteria and removing any duplicates, we are left with a total of 92 unique instances across the four Evaluations. We present results on these instances.

\paragraph{Incomplete score metric:} We use the same evaluation metric as the MaxSAT Evaluations, the \emph{incomplete score}. Let $c_{i}^j$ be the number of unsatisfied clauses achieved by the $i^{th}$ solver on the $j^{th}$ problem instance, and $c_{best}^j \leq c_{i}^j$ be the best known number of unsatisfied clauses for that instance. Then the incomplete score for solver $i$ on problem $j$ is defined as $s_i^j = (c_{best}^j + 1)/(c_{i}^j + 1)$. Note that $s_i^j \in (0, 1]$, with higher scores indicating better performance. As in the Evaluations, we rank solvers in decreasing order of their average incomplete score achieved on a set of instances at a wall clock time limit of 300 seconds. 

\paragraph{Baselines:} We compare RbmSAT to the solvers that participated in the MaxSAT Evaluations 2018-2021. The number of unsatisfied clauses achieved by a solver on each problem instance from a given Evaluation year is publicly available. We use this data to compute the average incomplete score of the participant solvers and RbmSAT on the selected instances. Since the set of solvers changes from year to year, the scores and the ranking of the solvers are computed separately for each year. The participant solvers use state-of-the art SAT solvers and stochastic local search algorithms and represent the current state-of-the-art for MaxSAT. Detailed descriptions can be found in \cite{maxsat2018, maxsat2019, maxsat2020, maxsat2021}.

\paragraph{Hardware configurations:}
Our best results showcase RbmSAT on each problem instance using a cluster of 64 TPUv4 chips. Solvers participating in Evaluations use conventional CPUs (Intel Xeon 2.4GHz CPU with 32 GB of RAM for the Incomplete Unweighted Track). These two hardware configurations are not easily comparable, so we run RbmSAT on CPU as well using the Evaluation machine configuration, which we denote as ``RbmSAT (CPU)'' in the results\footnote{Due to the complexity of porting our TPU-focused implementation to the competition execution environment, these results were obtained on a closely matched machine configuration, with a same-generation Intel Xeon CPU available in our own environment, and the same RAM limit.}.

\paragraph{Hyperparameter tuning:}
RbmSAT has several hyperparameters, such as the number of parallel chains to use\footnote{Note that more Markov chains is not always obviously better: running fewer chains may be preferable if it results in those chains making greater progress due to a lower time requirement per-step.} (i.e.\  batch size), frequency of applying the unit propagation-based heuristic to the samples, how to order variables in the heuristic, etc. We set the values of these hyperparameters using a grid search for each Evaluation year separately. We use problem instances from all Evaluations other than the one for which we are selecting the hyperparameters to evaluate the goodness of a hyperparameter choice.

\subsection{Results}
\label{subsec:results}
Table~\ref{tab:competition_average_incomplete_scores} summarizes the main results of the paper.
RbmSAT on 64 TPUs achieves a better average incomplete score at the 300 second time limit than Evaluation participants on instances from each of the four Evaluation years. RbmSAT (64 TPUs) achieves substantial improvements over the next best solver for the 2018 and 2019 Evaluations (24.9\% and 23.23\% better scores, respectively), with smaller improvements for the 2020 and 2021 Evaluations.
It achieves the lowest number of unsatisfied clauses on 35 out of the 92 instances across all four Evaluations. RbmSAT on CPU is also competitive, outperforming the other Evaluation participants on three out of the four years. This demonstrates that our stochastic search approach itself is powerful even without the compute throughput of a TPU cluster.
\begin{table}[]
    \centering
    \begin{footnotesize}
    \begin{sc}
    \begin{tabular}{l c}
    \toprule
    \multicolumn{2}{c}{MaxSAT 2018}\\
    \midrule
Solver & Average \\
    & incomplete \\
    & score\\
    \midrule
    \textbf{RbmSAT (64 TPUs)} & \textbf{0.947}\\
    RbmSAT (CPU) & 0.942\\
    SATLike & 0.758\\
    maxroster & 0.706\\
    SATLike-c & 0.699\\
    LinSBPS & 0.624\\
    Open-WBO-Inc-OBV & 0.620\\
    Open-WBO-Riss & 0.599\\
    Open-WBO-Gluc & 0.579\\
    Open-WBO-Inc-MCS & 0.559\\
    \midrule
    \midrule
    \multicolumn{2}{c}{MaxSAT 2019}\\
    \midrule
    \textbf{RbmSAT (64 TPUs)} & \textbf{0.968}\\
    RbmSAT (CPU) & 0.908\\
    SATLike & 0.785\\
    sls-mcs & 0.783\\
    sls-mcs-lsu & 0.782\\
    Loandra & 0.762\\
    LinSBPS2018 & 0.745\\
    Open-WBO-g & 0.712\\
    Open-WBO-ms & 0.709\\
\midrule
    \midrule
    \multicolumn{2}{c}{MaxSAT 2020}\\
    \midrule
\textbf{RbmSAT (64 TPUs)} & \textbf{0.930}\\
    RbmSAT (CPU) & 0.884\\
    sls-mcs & 0.866\\
    SATLike-c & 0.865\\
    sls-lsu & 0.865\\
    StableResolver & 0.863\\
    Loandra & 0.839\\
    TT-Open-WBO & 0.800\\
    \midrule
    \midrule
    \multicolumn{2}{c}{MaxSAT 2021}\\
    \midrule
    \textbf{RbmSAT (64 TPUs)} & \textbf{0.977}\\
    StableResolve & 0.975\\
    SATlike-c & 0.973\\
    Satlike-ck & 0.972\\
    TT-Open-WBO-Inc-21 & 0.973\\
    Loandra & 0.940\\
    RbmSAT (CPU) & 0.908\\
    Exact & 0.722\\
    \bottomrule
    \end{tabular}
    \end{sc}
    \end{footnotesize}
    \caption{\small Incomplete score at a time limit of 300s for RbmSAT and solvers that participated in the annual MaxSAT Evaluations 2018-2021, averaged over a subset of instances from the unweighted, incomplete solver track of the Evaluations that are selected to have maximum clause size $\leq 7$, number of variables $\leq 10000$, and number of clauses $\leq 100000$ to fit our implementation constraints (see section~\ref{sec:evaluation}). Higher incomplete score $\in (0,1]$ is better. ``RbmSAT (64 TPUs)'' runs on a cluster of 64 TPUs, while ``RbmSAT (CPU)'' runs on a CPU. Incomplete scores are computed using the number of unsatisfied clauses achieved by the participating solvers on the problem instances as reported at \url{https://maxsat-evaluations.github.io/} and by RbmSAT.}
    \label{tab:competition_average_incomplete_scores}
\end{table}

Table~\ref{tab:up_ablation} presents an ablation study that evaluates the effect of both the core RBM sampling component and the prioritized unit propagation heuristic.
Where the RBM is ablated and therefore conditional distribution variances are unavailable, a uniform random ranking prioritizes variables for unit propagation instead.
As these results demonstrate, both the RBM component and unit propagation contribute significantly to RbmSAT's performance.
For the 2018 and 2021 Evaluations, ablating the RBM seems to worsen results more than removing unit propagation, while for the other two Evaluations, the opposite is the case.
These results also demonstrate that while unit propagation is a powerful improvement operator for candidate assignments, repeated application of unit propagation to randomly chosen subsets is insufficient: a diverse set of strong candidates for improvement generated by stochastic search is critical, with the RBM performing better in this regard than candidates chosen uniformly at random.

\begin{table*}[t]
\centering
    \begin{sc}
    \centering
    \begin{tabular}{c cccc}
    \toprule
     & \multicolumn{4}{c}{Average incomplete score}\\
    Solver & MaxSAT 2018 & MaxSAT 2019 & MaxSAT 2020 & MaxSAT 2021\\
    \midrule
    RbmSAT & 0.947 & 0.968 & 0.930 & 0.977\\
    RbmSAT - UP & 0.938 & 0.842 & 0.782 & 0.840\\
    RbmSAT - Gibbs & 0.855 & 0.899  & 0.842 & 0.785\\
    UP only & 0.357 & 0.290 & 0.222 & 0.082\\
\bottomrule
    \end{tabular}
    \end{sc}
\caption{\small Ablation results for RbmSAT on 64 TPUs. ``RbmSAT - UP'' removes unit propagation, ``RbmSAT - Gibbs'' replaces Gibbs sampling with uniform random sampling, and ``UP only'' removes sampling and uses only unit propagation with a prioritization order chosen uniformly randomly.}
    \label{tab:up_ablation}
\end{table*}

\begin{figure}
    \centering
    \includegraphics[width=0.48\textwidth]{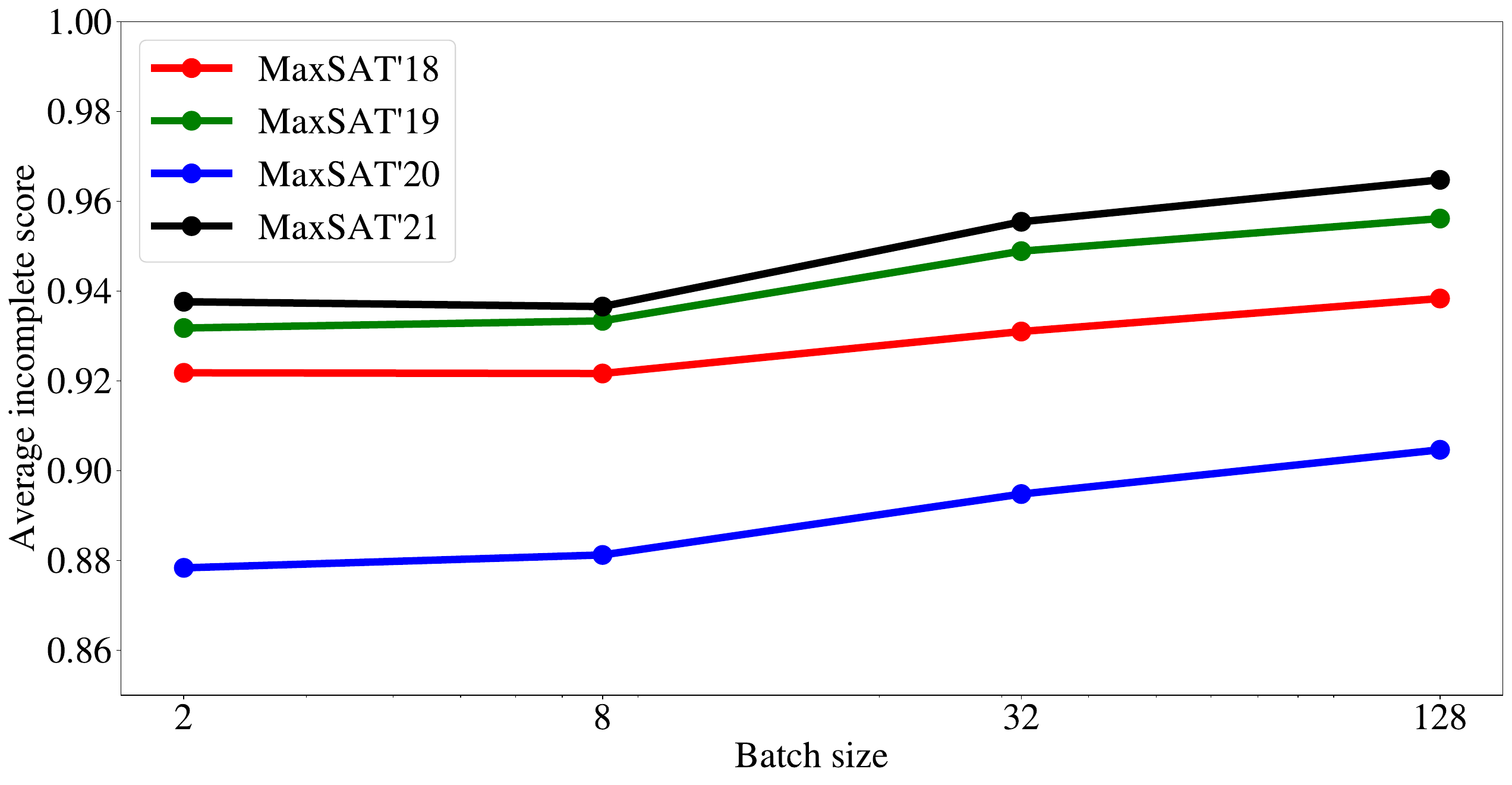}
\hfill
    \includegraphics[width=0.48\textwidth]{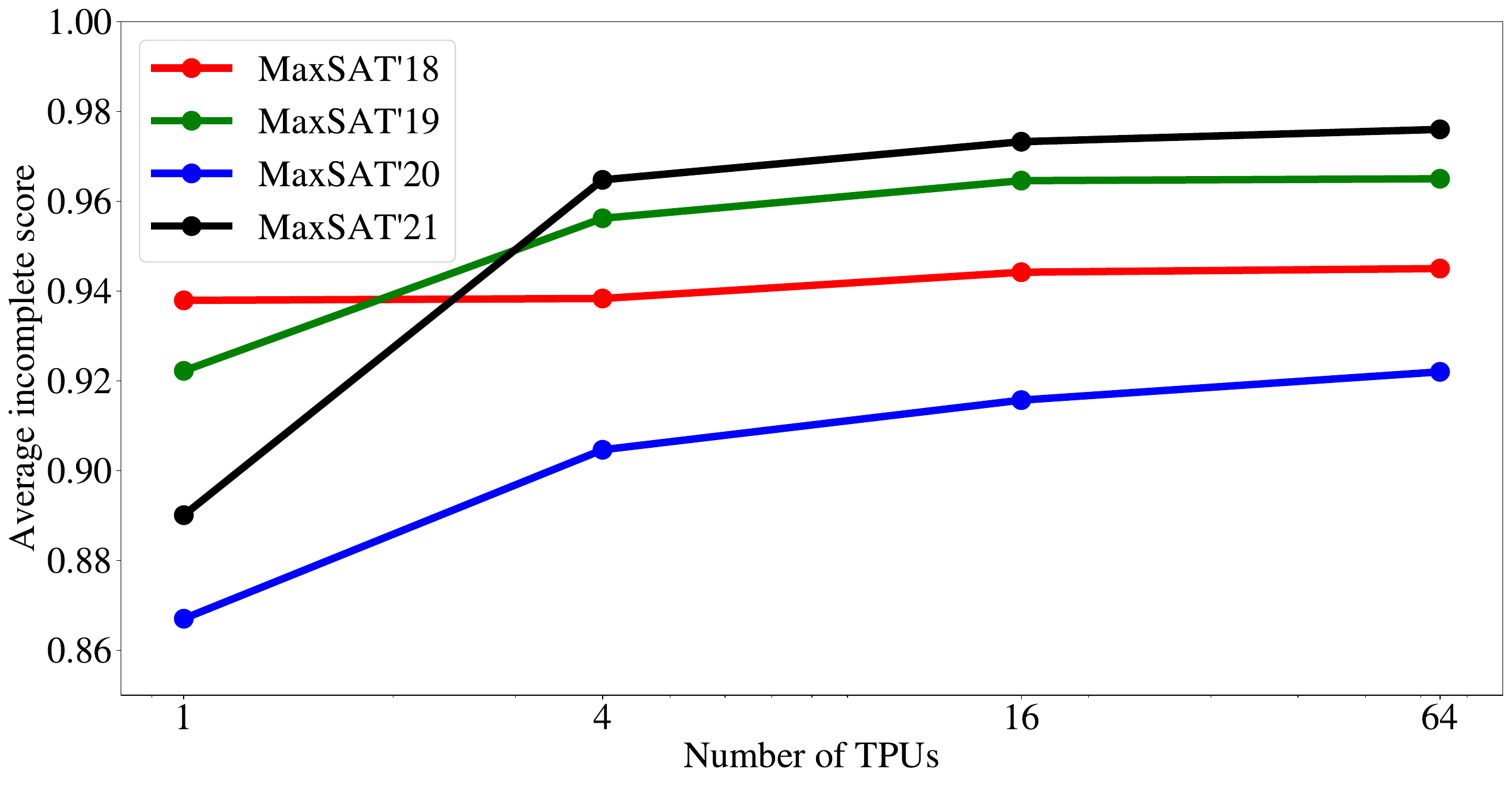}
\caption{\small Average incomplete score achieved by RbmSAT for various batch sizes (number of parallel Markov chains) per TPU device (left) and number of TPUs (right).}
    \label{fig:batchsize_ntpus_scaling}
\end{figure}
\begin{figure}[h!]
    \centering
    \includegraphics[width=0.46\textwidth]{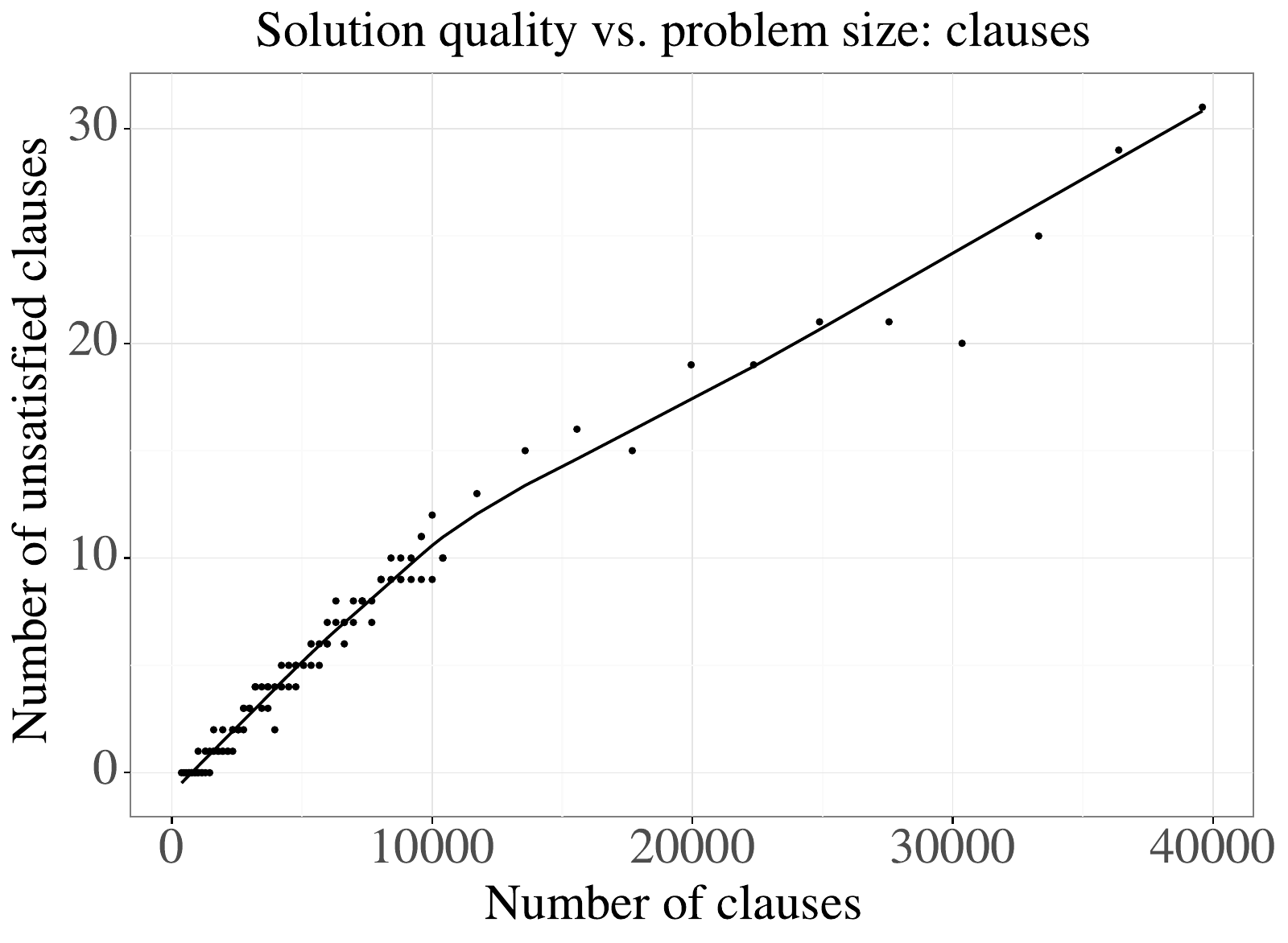}
\includegraphics[width=0.46\textwidth]{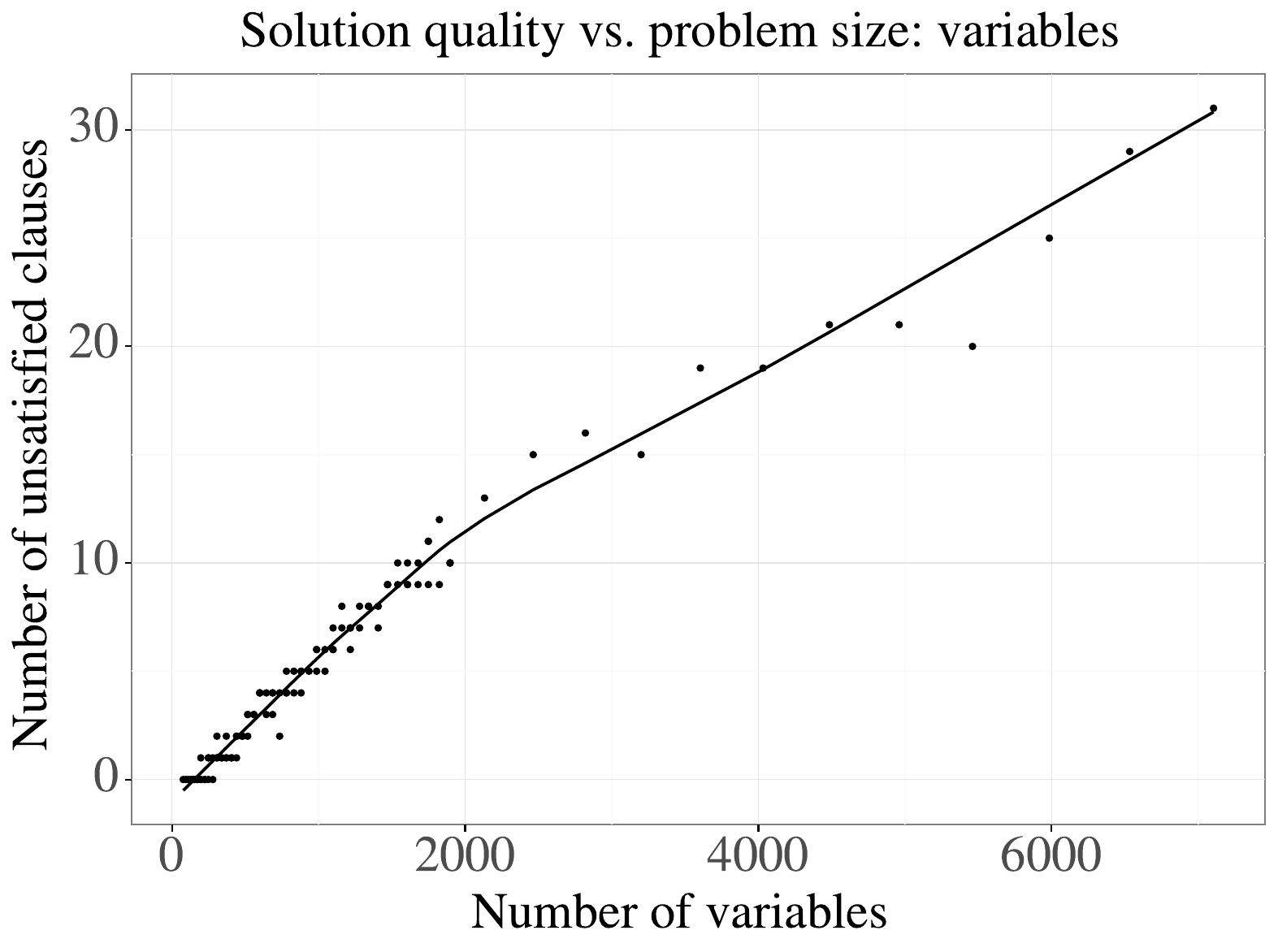}
    \caption{\small Number of unsatisfied clauses after 300 seconds on factoring instances as a function of the number of clauses (left) and number of variables.}
    \label{fig:problem_scaling}
\end{figure}
Figure~\ref{fig:batchsize_ntpus_scaling} shows the effect of increasing the number of parallel Markov chains (i.e., the batch size used during sampling) and the number of TPUs on the average incomplete score. As expected, the results improve with more chains and more TPUs. Unsurprisingly, the graphs show diminishing returns for both, particularly for the number of TPUs (which corresponds to larger jumps in the number of chains). 

Figure~\ref{fig:problem_scaling} illustrates preliminary results on the scaling behaviour of RbmSAT as a function of the problem size in terms of the number of variables and clauses in the problem. 
The curve illustrates performance of 64 TPUv2 chips with a 300 second wall clock time budget on an arbitrarily chosen subset of the ``long'' factoring instances encoded as SAT problems, as proposed in \cite{mosca2019factoring}\footnote{These instances can be downloaded at \url{https://github.com/sebastianv89/factoring-sat/tree/master/instances}} where the product to be factored ranges in size from 6 to 100 bits. The number of unsatisfied clauses has a roughly linear relationship with the number of variables or clauses. However, the problems used here are relatively small; further study on larger instances is needed to better understand RbmSAT's scaling behavior.

  \section{Conclusions and Future Work}
We have derived a correspondence between RBMs and MaxSAT problems and proposed RbmSAT, a highly parallelizable algorithm that can exploit the compute power of neural network accelerators for MaxSAT. We have demonstrated its efficacy on instances drawn from four MaxSAT Evaluations, and showcased the scaling benefits of our approach. Our results, including a baseline which demonstrates lower but nonetheless surprisingly strong performance simply by filtering high-throughput randomly generated assignments and improving them with unit propagation, suggest that our contribution will be far from the last word on hardware-accelerated stochastic search for MaxSAT and related problems. Just as deep learning workloads benefited from advances in GPU technology driven by commodity video gaming hardware, our RBM construction of SAT is well positioned to exploit the wide availability of accelerated matrix multiplication hardware primarily built for training and inference in deterministic deep neural networks.

One future direction is to apply RbmSAT to SAT and \#SAT. Though RbmSAT is most naturally thought of as a MaxSAT search algorithm given its construction of a distribution with monotonically increasing probability in the number of satisfied clauses, RbmSAT can identify fully satisfying assignments. Improvements which encourage assignment diversity may adapt RbmSAT more readily to SAT. \#SAT is the problem of finding the total number of fully satisfying assignments, which can be posed as the problem of computing the partition function of the formula-RBM at the zero temperature limit when all fully satisfying assignments have equal probability while the rest have zero probability. Techniques such as Annealed Importance Sampling (AIS) \cite{neal1998ais}, which has already been explored for the estimation of RBM partition functions~\cite{salahutdinov2008aisrbm}, can potentially be applied to finding approximate solutions for \#SAT.

Our method involves running an ensemble of RBMs trained at different temperatures, and could in principle be extended to use parallel tempering~\citep{swendsen1986replica,desjardins2010parallel,cho2010parallel}, which simulates an ensemble of systems at different temperatures and periodically proposes swaps between particles in adjacent replicas. This procedure allows particles to mix between disparate modes at higher temperatures, while particles at lower temperatures explore the fine structure of the energy landscape within modes, with exchanges allowing a single particle to mix more freely on finite time horizons than Gibbs dynamics would ordinarily permit.

 \section*{Acknowledgements}
We would like to thank Alex Davies for detailed technical feedback on the manuscript as well as Oriol Vinyals, Guillaume Desjardins and Geoffrey Hinton for helpful research discussions, Aidan Clark for early engineering discussions, Jonas Adler, Andy Brock, John Jumper and Sander Dieleman for valuable advice regarding TPUs, and Richard Tanburn for debugging support. 
We would also like to acknowledge the foundational work of the JAX and XLA teams, in particular Lena Martens, Blake Hechtman and Peter Hawkins, as well as all contributors to the DeepMind JAX ecosystem~\cite{deepmind2020jax}.

\bibliographystyle{plain}
\bibliography{references}

\begin{thebibliography}{10}

\bibitem{cloud-tpu-description}
{Cloud TPU System Architecture}.
\newblock \url{https://cloud.google.com/tpu/docs/system-architecture-tpu-vm}.
\newblock Accessed: 2022-05-19.

\bibitem{maxsat-eval}
{MaxSAT Evaluations}.
\newblock \url{https://maxsat-evaluations.github.io/}.
\newblock Accessed: 2022-05-19.

\bibitem{nvidia2022h100}
Nvidia h100 tensor core gpu architecture.
\newblock \url{https://resources.nvidia.com/en-us-tensor-core}.
\newblock Accessed: 2022-10-13.

\bibitem{sat-comp}
{SAT Competitions}.
\newblock \url{http://www.satcompetition.org/}.
\newblock Accessed: 2022-05-19.

\bibitem{xla2017}
Xla: Optimizing compiler for machine learning, 2017.

\bibitem{amizadeh2019LearningTS}
Saeed Amizadeh, Sergiy Matusevych, and Markus Weimer.
\newblock Learning to solve circuit-sat: An unsupervised differentiable
  approach.
\newblock In {\em ICLR}, 2019.

\bibitem{deepmind2020jax}
Igor Babuschkin, Kate Baumli, Alison Bell, Surya Bhupatiraju, Jake Bruce, Peter
  Buchlovsky, David Budden, Trevor Cai, Aidan Clark, Ivo Danihelka, Antoine
  Dedieu, Claudio Fantacci, Jonathan Godwin, Chris Jones, Ross Hemsley, Tom
  Hennigan, Matteo Hessel, Shaobo Hou, Steven Kapturowski, Thomas Keck, Iurii
  Kemaev, Michael King, Markus Kunesch, Lena Martens, Hamza Merzic, Vladimir
  Mikulik, Tamara Norman, George Papamakarios, John Quan, Roman Ring, Francisco
  Ruiz, Alvaro Sanchez, Rosalia Schneider, Eren Sezener, Stephen Spencer,
  Srivatsan Srinivasan, Wojciech Stokowiec, Luyu Wang, Guangyao Zhou, and Fabio
  Viola.
\newblock The {D}eep{M}ind {JAX} {E}cosystem, 2020.

\bibitem{maxsat2020}
Fahiem Bacchus, Jeremias Berg, Matti Järvisalo, and Ruben Martins.
\newblock Maxsat evaluation 2020: Solver and benchmark descriptions.
\newblock {\em University of Helsinki Department of Computer Science Report
  Series B}, B-2020-2, 2020.

\bibitem{maxsat2021}
Fahiem Bacchus, Jeremias Berg, Matti Järvisalo, and Ruben Martins.
\newblock Maxsat evaluation 2021: Solver and benchmark descriptions.
\newblock {\em University of Helsinki Department of Computer Science Report
  Series B}, B-2021-2, 2021.

\bibitem{bacchus2021MaxSAT}
Fahiem Bacchus, Matti J{\"a}rvisalo, and Ruben Martins.
\newblock {\em Maximum Satisfiability}, pages 929 -- 991.
\newblock Frontiers in Artificial Intelligence and Applications. IOS PRESS,
  Netherlands, 2 edition, 2021.

\bibitem{maxsat2018}
Fahiem Bacchus, Matti Järvisalo, and Ruben Martins.
\newblock Maxsat evaluation 2018: Solver and benchmark descriptions.
\newblock {\em University of Helsinki Department of Computer Science Series of
  Publications B}, B-2018-2, 2018.

\bibitem{maxsat2019}
Fahiem Bacchus, Matti Järvisalo, and Ruben Martins.
\newblock Maxsat evaluation 2019: Solver and benchmark descriptions.
\newblock {\em University of Helsinki Department of Computer Science Report
  Series B}, B-2019-2, 2019.

\bibitem{biere2009sat-handbook}
A.~Biere, A.~Biere, M.~Heule, H.~van Maaren, and T.~Walsh.
\newblock {\em Handbook of Satisfiability: Volume 185 Frontiers in Artificial
  Intelligence and Applications}.
\newblock IOS Press, NLD, 2009.

\bibitem{biere2021handbook}
A.~Biere, M.~Heule, and H.~van Maaren.
\newblock {\em Handbook of Satisfiability: Second Edition}.
\newblock Frontiers in Artificial Intelligence and Applications. IOS Press,
  2021.

\bibitem{jax2018github}
James Bradbury, Roy Frostig, Peter Hawkins, Matthew~James Johnson, Chris Leary,
  Dougal Maclaurin, George Necula, Adam Paszke, Jake Vander{P}las, Skye
  Wanderman-{M}ilne, and Qiao Zhang.
\newblock {JAX}: composable transformations of {P}ython+{N}um{P}y programs,
  2018.

\bibitem{braunstein2005surveyprop}
Alfredo Braunstein, Marc Mezard, and Riccardo Zecchina.
\newblock Survey propagation: An algorithm for satisfiability.
\newblock {\em Random Struct. Algorithms}, 27:201--226, 09 2005.

\bibitem{cho2010parallel}
KyungHyun Cho, Tapani Raiko, and Alexander Ilin.
\newblock Parallel tempering is efficient for learning restricted boltzmann
  machines.
\newblock In {\em The 2010 international joint conference on neural networks
  (ijcnn)}, pages 1--8. IEEE, 2010.

\bibitem{cook2019isinggpu}
Chase Cook, Hengyang Zhao, Takashi Sato, Masayuki Hiromoto, and Sheldon X.~D.
  Tan.
\newblock Gpu based parallel ising computing for combinatorial optimization
  problems in vlsi physical design, 2019.

\bibitem{Cook:1971}
Stephen~A. Cook.
\newblock The complexity of theorem-proving procedures.
\newblock In {\em Proceedings of the Third Annual ACM Symposium on Theory of
  Computing}, STOC '71, pages 151--158, New York, NY, USA, 1971. ACM.

\bibitem{anjou1993bmsat}
A.~d'Anjou, M.~Grana, F.J. Torrealdea, and M.C. Hernandez.
\newblock Solving satisfiability via boltzmann machines.
\newblock {\em IEEE Transactions on Pattern Analysis and Machine Intelligence},
  15(5):514--521, 1993.

\bibitem{desjardins2010parallel}
Guillaume Desjardins, Aaron Courville, Yoshua Bengio, Pascal Vincent, Olivier
  Delalleau, et~al.
\newblock Parallel tempering for training of restricted boltzmann machines.
\newblock In {\em Proceedings of the thirteenth international conference on
  artificial intelligence and statistics}, pages 145--152. MIT Press Cambridge,
  MA, 2010.

\bibitem{fichte2020timeleapSAT}
Johannes~K. Fichte, Markus Hecher, and Stefan Szeider.
\newblock A time leap challenge for sat-solving.
\newblock In {\em Principles and Practice of Constraint Programming: 26th
  International Conference, CP 2020, Louvain-La-Neuve, Belgium, September
  7–11, 2020, Proceedings}, page 267–285, Berlin, Heidelberg, 2020.
  Springer-Verlag.

\bibitem{freund1991unsupervised}
Yoav Freund and David Haussler.
\newblock Unsupervised learning of distributions on binary vectors using two
  layer networks.
\newblock In J.~Moody, S.~Hanson, and R.~P. Lippmann, editors, {\em Advances in
  Neural Information Processing Systems}, volume~4. Morgan-Kaufmann, 1992.

\bibitem{gu2019rbms}
Linyan Gu, Jianfeng Huang, and Lihua Yang.
\newblock On the representational power of restricted boltzmann machines for
  symmetric functions and boolean functions.
\newblock {\em IEEE Transactions on Neural Networks and Learning Systems},
  30:1335--1347, 2019.

\bibitem{heule2012cube_and_conquer}
Marijn J.~H. Heule, Oliver Kullmann, Siert Wieringa, and Armin Biere.
\newblock Cube and conquer: Guiding cdcl sat solvers by lookaheads.
\newblock In Kerstin Eder, Jo{\~a}o Louren{\c{c}}o, and Onn Shehory, editors,
  {\em Hardware and Software: Verification and Testing}, pages 50--65, Berlin,
  Heidelberg, 2012. Springer Berlin Heidelberg.

\bibitem{hinton2002training}
Geoffrey~E Hinton.
\newblock Training products of experts by minimizing contrastive divergence.
\newblock {\em Neural Computation}, 14(8):1771--1800, 2002.

\bibitem{hinton1986learning}
Geoffrey~E Hinton, Terrence~J Sejnowski, et~al.
\newblock Learning and relearning in boltzmann machines.
\newblock In {\em Parallel distributed processing: Explorations in the
  microstructure of cognition}, volume~1, pages 282--317. MIT Press, 1986.

\bibitem{hoos2000LSforSAT}
Holger Hoos and Thomas Stützle.
\newblock Local search algorithms for sat: An empirical evaluation.
\newblock {\em Journal of Automated Reasoning}, 24:421--481, 01 2000.

\bibitem{hopfieldtank1985tsp}
John Hopfield and D~Tank.
\newblock Neural computation of decisions in optimization problems.
\newblock {\em Biological cybernetics}, 52:141--52, 02 1985.

\bibitem{jouppi2021tenlessons}
Norman~P. Jouppi, Doe~Hyun Yoon, Matthew Ashcraft, Mark Gottscho, Thomas~B.
  Jablin, George Kurian, James Laudon, Sheng Li, Peter Ma, Xiaoyu Ma, Thomas
  Norrie, Nishant Patil, Sushma Prasad, Cliff Young, Zongwei Zhou, and David
  Patterson.
\newblock {\em Ten Lessons from Three Generations Shaped Google's TPUv4i}, page
  1–14.
\newblock IEEE Press, 2021.

\bibitem{jouppi2017tpus}
Norman~P. Jouppi, Cliff Young, Nishant Patil, David Patterson, Gaurav Agrawal,
  Raminder Bajwa, Sarah Bates, Suresh Bhatia, Nan Boden, Al~Borchers, Rick
  Boyle, Pierre-luc Cantin, Clifford Chao, Chris Clark, Jeremy Coriell, Mike
  Daley, Matt Dau, Jeffrey Dean, Ben Gelb, Tara~Vazir Ghaemmaghami, Rajendra
  Gottipati, William Gulland, Robert Hagmann, C.~Richard Ho, Doug Hogberg, John
  Hu, Robert Hundt, Dan Hurt, Julian Ibarz, Aaron Jaffey, Alek Jaworski,
  Alexander Kaplan, Harshit Khaitan, Daniel Killebrew, Andy Koch, Naveen Kumar,
  Steve Lacy, James Laudon, James Law, Diemthu Le, Chris Leary, Zhuyuan Liu,
  Kyle Lucke, Alan Lundin, Gordon MacKean, Adriana Maggiore, Maire Mahony,
  Kieran Miller, Rahul Nagarajan, Ravi Narayanaswami, Ray Ni, Kathy Nix, Thomas
  Norrie, Mark Omernick, Narayana Penukonda, Andy Phelps, Jonathan Ross, Matt
  Ross, Amir Salek, Emad Samadiani, Chris Severn, Gregory Sizikov, Matthew
  Snelham, Jed Souter, Dan Steinberg, Andy Swing, Mercedes Tan, Gregory
  Thorson, Bo~Tian, Horia Toma, Erick Tuttle, Vijay Vasudevan, Richard Walter,
  Walter Wang, Eric Wilcox, and Doe~Hyun Yoon.
\newblock In-datacenter performance analysis of a tensor processing unit.
\newblock In {\em Proceedings of the 44th Annual International Symposium on
  Computer Architecture}, ISCA '17, page 1–12, New York, NY, USA, 2017.
  Association for Computing Machinery.

\bibitem{kautz2021incomplete}
Henry~A. Kautz, Ashish Sabharwal, and Bart Selman.
\newblock Incomplete algorithms.
\newblock In Armin Biere, Marijn Heule, Hans van Maaren, and Toby Walsh,
  editors, {\em Handbook of Satisfiability - Second Edition}, volume 336 of
  {\em Frontiers in Artificial Intelligence and Applications}, pages 213--232.
  {IOS} Press, 2021.

\bibitem{kingma2014adam}
Diederik Kingma and Jimmy Ba.
\newblock Adam: {A} method for stochastic optimization.
\newblock arXiv preprint arXiv:1412.6980, 2014.

\bibitem{kotthoff2018quantifying}
Lars Kotthoff, Alexandre Fr\'{e}chette, Tomasz Michalak, Talal Rahwan,
  Holger~H. Hoos, and Kevin Leyton-Brown.
\newblock Quantifying algorithmic improvements over time.
\newblock In {\em Proceedings of the 27th International Joint Conference on
  Artificial Intelligence}, IJCAI'18, page 5165–5171. AAAI Press, 2018.

\bibitem{leroux2008representational}
Nicolas Le~Roux and Yoshua Bengio.
\newblock Representational power of restricted boltzmann machines and deep
  belief networks.
\newblock {\em Neural Computation}, 20(6):1631--1649, 2008.

\bibitem{li2018guidedTreeSearch}
Zhuwen Li, Qifeng Chen, and Vladlen Koltun.
\newblock Combinatorial optimization with graph convolutional networks and
  guided tree search.
\newblock In S.~Bengio, H.~Wallach, H.~Larochelle, K.~Grauman, N.~Cesa-Bianchi,
  and R.~Garnett, editors, {\em Advances in Neural Information Processing
  Systems}, volume~31. Curran Associates, Inc., 2018.

\bibitem{lucas2014IsingforNP}
Andrew {Lucas}.
\newblock {Ising formulations of many NP problems}.
\newblock {\em Frontiers in Physics}, 2:5, February 2014.

\bibitem{marques-silva2008SATapplications}
Joao Marques-Silva.
\newblock Practical applications of boolean satisfiability.
\newblock In {\em 2008 9th International Workshop on Discrete Event Systems},
  pages 74--80, 2008.

\bibitem{MarquesSilva2009ConflictDrivenCL}
Joao Marques-Silva, In{\^e}s Lynce, and Sharad Malik.
\newblock Conflict-driven clause learning sat solvers.
\newblock In {\em Handbook of Satisfiability}, 2009.

\bibitem{martens2013rbms}
James Martens, Arkadev Chattopadhya, Toni Pitassi, and Richard Zemel.
\newblock On the representational efficiency of restricted boltzmann machines.
\newblock In C.~J.~C. Burges, L.~Bottou, M.~Welling, Z.~Ghahramani, and K.~Q.
  Weinberger, editors, {\em Advances in Neural Information Processing Systems},
  volume~26. Curran Associates, Inc., 2013.

\bibitem{montufar2011refinements}
Guido Montufar and Nihat Ay.
\newblock {Refinements of Universal Approximation Results for Deep Belief
  Networks and Restricted Boltzmann Machines}.
\newblock {\em Neural Computation}, 23(5):1306--1319, 05 2011.

\bibitem{mosca2019factoring}
Michele Mosca and Sebastian~R Verschoor.
\newblock Factoring semi-primes with (quantum) sat-solvers.
\newblock {\em arXiv preprint arXiv:1902.01448}, 2019.

\bibitem{neal1998ais}
Radford~M. Neal.
\newblock Annealed importance sampling.
\newblock {\em Statistics and Computing}, 11, 2001.

\bibitem{patel2019combining}
Saavan Patel and Sayeef Salahuddin.
\newblock Combining learned representations for combinatorial optimization,
  2019.

\bibitem{salahutdinov2008aisrbm}
Ruslan Salakhutdinov and Iain Murray.
\newblock On the quantitative analysis of deep belief networks.
\newblock In {\em Proceedings of the 25th International Conference on Machine
  Learning}, ICML '08, page 872–879, New York, NY, USA, 2008. Association for
  Computing Machinery.

\bibitem{selsam2019guiding}
Daniel Selsam and Nikolaj Bj{\o}rner.
\newblock Guiding high-performance sat solvers with unsat-core predictions.
\newblock In Mikol{\'a}{\v{s}} Janota and In{\^e}s Lynce, editors, {\em Theory
  and Applications of Satisfiability Testing -- SAT 2019}, pages 336--353,
  Cham, 2019. Springer International Publishing.

\bibitem{selsam2019NeuroSAT}
Daniel Selsam, Matthew Lamm, Benedikt B{\"{u}}nz, Percy Liang, Leonardo
  de~Moura, and David~L. Dill.
\newblock Learning a {SAT} solver from single-bit supervision.
\newblock In {\em 7th International Conference on Learning Representations,
  {ICLR} 2019, New Orleans, LA, USA, May 6-9, 2019}, 2019.

\bibitem{si2019prioritized}
Xujie Si, Yujia Li, Vinod Nair, and Felix Gimeno.
\newblock Prioritized unit propagation with periodic resetting is (almost) all
  you need for random {SAT} solving.
\newblock {\em CoRR}, abs/1912.05906, 2019.

\bibitem{smolensky1986information}
P~Smolensky.
\newblock Information processing in dynamical systems: foundations of harmony
  theory.
\newblock In {\em Parallel distributed processing: explorations in the
  microstructure of cognition}, volume~1, pages 194--281. MIT Press, 1986.

\bibitem{SOHANGHPURWALA2017hwsat}
Ali~Asgar Sohanghpurwala, Mohamed~W. Hassan, and Peter Athanas.
\newblock Hardware accelerated sat solvers—a survey.
\newblock {\em Journal of Parallel and Distributed Computing}, 106:170--184,
  2017.

\bibitem{sutskever2010convergence}
Ilya Sutskever and Tijmen Tieleman.
\newblock On the convergence properties of contrastive divergence.
\newblock In Yee~Whye Teh and Mike Titterington, editors, {\em Proceedings of
  the Thirteenth International Conference on Artificial Intelligence and
  Statistics}, volume~9 of {\em Proceedings of Machine Learning Research},
  pages 789--795, Chia Laguna Resort, Sardinia, Italy, 13--15 May 2010. PMLR.

\bibitem{swendsen1986replica}
Robert~H Swendsen and Jian-Sheng Wang.
\newblock Replica monte carlo simulation of spin-glasses.
\newblock {\em Physical review letters}, 57(21):2607, 1986.

\bibitem{yang2019isingtpu}
Kun Yang, Yi-Fan Chen, Georgios Roumpos, Chris Colby, and John Anderson.
\newblock High performance monte carlo simulation of ising model on tpu
  clusters.
\newblock In {\em Proceedings of the International Conference for High
  Performance Computing, Networking, Storage and Analysis}, SC '19, New York,
  NY, USA, 2019. Association for Computing Machinery.

\bibitem{yolcu2019learnLS}
Emre Yolcu and Barnabas Poczos.
\newblock Learning local search heuristics for boolean satisfiability.
\newblock In {\em Advances in Neural Information Processing Systems},
  volume~32. Curran Associates, Inc., 2019.

\end{thebibliography}

\clearpage
\section*{Appendix}
\label{sec:appendix}
\subsection{JAX implementation of RbmSAT}
\label{subsec:algorithm}
\begin{figure}[h]
\label{fig:jaximpl}
\centering
\tiny{
\begin{Verbatim}[commandchars=\\\{\},numbers=left,firstnumber=1,stepnumber=1,xleftmargin=6mm]
\PY{c+c1}{\PYZsh{} Copyright 2023 DeepMind Technologies Limited.}
\PY{c+c1}{\PYZsh{} SPDX\PYZhy{}License\PYZhy{}Identifier: Apache\PYZhy{}2.0}

\PY{k+kn}{from} \PY{n+nn}{jax} \PY{k+kn}{import} \PY{n}{nn}
\PY{k+kn}{import} \PY{n+nn}{jax}\PY{n+nn}{.}\PY{n+nn}{numpy} \PY{k}{as} \PY{n+nn}{jnp}
\PY{k+kn}{from} \PY{n+nn}{jax}\PY{n+nn}{.}\PY{n+nn}{random} \PY{k+kn}{import} \PY{n}{PRNGKey}\PY{p}{,} \PY{n}{bernoulli}\PY{p}{,} \PY{n}{fold\PYZus{}in}\PY{p}{,} \PY{n}{split}


\PY{k}{def} \PY{n+nf}{gather\PYZus{}and\PYZus{}count}\PY{p}{(}\PY{n}{v}\PY{p}{,} \PY{n}{Q}\PY{p}{,} \PY{n}{polarity}\PY{p}{)}\PY{p}{:}
  \PY{n}{c} \PY{o}{=} \PY{n}{jnp}\PY{o}{.}\PY{n}{einsum}\PY{p}{(}\PY{l+s+s1}{\PYZsq{}}\PY{l+s+s1}{bv,kcv\PYZhy{}\PYZgt{}bkc}\PY{l+s+s1}{\PYZsq{}}\PY{p}{,} \PY{n}{v}\PY{p}{,} \PY{n}{Q}\PY{p}{)}
  \PY{k}{return} \PY{n}{c}\PY{p}{,} \PY{p}{(}\PY{p}{(}\PY{n}{polarity} \PY{o}{\PYZgt{}} \PY{l+m+mi}{0}\PY{p}{)} \PY{o}{*} \PY{n}{c}\PY{p}{)}\PY{o}{.}\PY{n}{any}\PY{p}{(}\PY{n}{axis}\PY{o}{=}\PY{o}{\PYZhy{}}\PY{l+m+mi}{2}\PY{p}{)}\PY{o}{.}\PY{n}{sum}\PY{p}{(}\PY{n}{axis}\PY{o}{=}\PY{o}{\PYZhy{}}\PY{l+m+mi}{1}\PY{p}{)}


\PY{k}{def} \PY{n+nf}{rbmsat}\PY{p}{(}\PY{n}{W\PYZus{}c}\PY{p}{,} \PY{n}{b\PYZus{}c}\PY{p}{,} \PY{n}{T}\PY{p}{,} \PY{n}{B}\PY{p}{,} \PY{n}{N}\PY{p}{,} \PY{n}{upp}\PY{p}{,} \PY{n}{upw}\PY{p}{,} \PY{n}{seed}\PY{p}{,} \PY{n}{\ensuremath{\alpha}}\PY{p}{)}\PY{p}{:}
  \PY{n}{s\PYZus{}max} \PY{o}{=} \PY{n}{jnp}\PY{o}{.}\PY{n}{zeros}\PY{p}{(}\PY{p}{(}\PY{n}{B}\PY{p}{,}\PY{p}{)}\PY{p}{)}
  \PY{n}{init\PYZus{}key}\PY{p}{,} \PY{n}{v\PYZus{}key}\PY{p}{,} \PY{n}{h\PYZus{}key} \PY{o}{=} \PY{n}{split}\PY{p}{(}\PY{n}{PRNGKey}\PY{p}{(}\PY{n}{seed}\PY{p}{)}\PY{p}{,} \PY{l+m+mi}{3}\PY{p}{)}
  \PY{n}{v} \PY{o}{=} \PY{n}{bernoulli}\PY{p}{(}\PY{n}{init\PYZus{}key}\PY{p}{,} \PY{n}{shape}\PY{o}{=}\PY{p}{(}\PY{n}{B}\PY{p}{,} \PY{n}{N}\PY{p}{)}\PY{p}{)}
  \PY{n}{\ensuremath{\nu}} \PY{o}{=} \PY{l+m+mf}{0.25} \PY{o}{*} \PY{n}{jnp}\PY{o}{.}\PY{n}{ones\PYZus{}like}\PY{p}{(}\PY{n}{v}\PY{p}{)}
  \PY{n}{polarity} \PY{o}{=} \PY{n}{jnp}\PY{o}{.}\PY{n}{sign}\PY{p}{(}\PY{n}{T}\PY{p}{)}  \PY{c+c1}{\PYZsh{} strictly \PYZob{}1, \PYZhy{}1\PYZcb{}, not 0}
  \PY{n}{Q} \PY{o}{=} \PY{n}{nn}\PY{o}{.}\PY{n}{one\PYZus{}hot}\PY{p}{(}\PY{n+nb}{abs}\PY{p}{(}\PY{n}{T}\PY{p}{)} \PY{o}{\PYZhy{}} \PY{l+m+mi}{1}\PY{p}{,} \PY{n}{N}\PY{p}{)}  \PY{c+c1}{\PYZsh{} indices in T are 1\PYZhy{}based}
  \PY{n}{W} \PY{o}{=} \PY{n}{polarity}\PY{p}{[}\PY{p}{:}\PY{p}{,} \PY{n}{jnp}\PY{o}{.}\PY{n}{newaxis}\PY{p}{,} \PY{p}{:}\PY{p}{]} \PY{o}{*} \PY{n}{W\PYZus{}c}\PY{p}{[}\PY{p}{:}\PY{p}{,} \PY{p}{:}\PY{p}{,} \PY{n}{jnp}\PY{o}{.}\PY{n}{newaxis}\PY{p}{]}
  \PY{n}{b} \PY{o}{=} \PY{n}{b\PYZus{}c}\PY{p}{[}\PY{p}{:}\PY{p}{,} \PY{n}{jnp}\PY{o}{.}\PY{n}{newaxis}\PY{p}{]} \PY{o}{+} \PY{n}{W\PYZus{}c}\PY{o}{.}\PY{n}{T} \PY{o}{@} \PY{p}{(}\PY{l+m+mi}{1} \PY{o}{\PYZhy{}} \PY{n}{polarity}\PY{p}{)} \PY{o}{/} \PY{l+m+mi}{2}
  \PY{n}{t}\PY{p}{,} \PY{n}{d} \PY{o}{=} \PY{l+m+mi}{1}\PY{p}{,} \PY{o}{\PYZhy{}}\PY{l+m+mi}{1}
  \PY{n}{step} \PY{o}{=} \PY{l+m+mi}{0}
  \PY{k}{while} \PY{n}{time\PYZus{}remaining}\PY{p}{(}\PY{p}{)}\PY{p}{:}
    \PY{n}{step} \PY{o}{+}\PY{o}{=} \PY{l+m+mi}{1}
    \PY{k}{if} \PY{n}{upp} \PY{o}{\PYZgt{}} \PY{l+m+mi}{0} \PY{o+ow}{and} \PY{n}{d} \PY{o}{==} \PY{l+m+mi}{0}\PY{p}{:}
      \PY{n}{v\PYZus{}u} \PY{o}{=} \PY{n}{jnp}\PY{o}{.}\PY{n}{vstack}\PY{p}{(}\PY{p}{[}\PY{n}{v}\PY{p}{,} \PY{n}{fetch\PYZus{}unit\PYZus{}prop\PYZus{}result}\PY{p}{(}\PY{p}{)}\PY{p}{]}\PY{p}{)}
      \PY{n}{\PYZus{}}\PY{p}{,} \PY{n}{s\PYZus{}u} \PY{o}{=} \PY{n}{gather\PYZus{}and\PYZus{}count}\PY{p}{(}\PY{n}{v\PYZus{}u}\PY{p}{,} \PY{n}{Q}\PY{p}{,} \PY{n}{polarity}\PY{p}{)}
      \PY{n}{ranks} \PY{o}{=} \PY{n}{jnp}\PY{o}{.}\PY{n}{argsort}\PY{p}{(}\PY{n}{s\PYZus{}u}\PY{p}{)}
      \PY{n}{v} \PY{o}{=} \PY{n}{v\PYZus{}u}\PY{p}{[}\PY{n}{ranks}\PY{p}{[}\PY{o}{\PYZhy{}}\PY{n}{B}\PY{p}{:}\PY{p}{]}\PY{p}{]}
      \PY{n}{\ensuremath{\nu}} \PY{o}{=} \PY{n}{jnp}\PY{o}{.}\PY{n}{vstack}\PY{p}{(}\PY{p}{[}\PY{n}{\ensuremath{\nu}}\PY{p}{,} \PY{n}{\ensuremath{\nu}}\PY{p}{]}\PY{p}{)}\PY{p}{[}\PY{n}{ranks}\PY{p}{[}\PY{o}{\PYZhy{}}\PY{n}{B}\PY{p}{:}\PY{p}{]}\PY{p}{]}
    \PY{k}{if} \PY{n}{upp} \PY{o}{\PYZgt{}} \PY{l+m+mi}{0} \PY{o+ow}{and} \PY{n}{t} \PY{o}{\PYZpc{}} \PY{n}{upp} \PY{o}{==} \PY{l+m+mi}{0}\PY{p}{:}
      \PY{n}{d} \PY{o}{=} \PY{n}{upw}
      \PY{n}{send\PYZus{}for\PYZus{}unit\PYZus{}propagation}\PY{p}{(}\PY{n}{v}\PY{p}{,} \PY{n}{\ensuremath{\nu}}\PY{p}{)}
    \PY{n}{c}\PY{p}{,} \PY{n}{s} \PY{o}{=} \PY{n}{gather\PYZus{}and\PYZus{}count}\PY{p}{(}\PY{n}{v}\PY{p}{,} \PY{n}{Q}\PY{p}{,} \PY{n}{polarity}\PY{p}{)}
    \PY{n}{s\PYZus{}max} \PY{o}{=} \PY{n}{jnp}\PY{o}{.}\PY{n}{maximum}\PY{p}{(}\PY{n}{s\PYZus{}max}\PY{p}{,} \PY{n}{s}\PY{p}{)}
    \PY{n}{h\PYZus{}logits} \PY{o}{=} \PY{n}{b} \PY{o}{+} \PY{n}{jnp}\PY{o}{.}\PY{n}{einsum}\PY{p}{(}\PY{l+s+s1}{\PYZsq{}}\PY{l+s+s1}{bkc,khc\PYZhy{}\PYZgt{}bhc}\PY{l+s+s1}{\PYZsq{}}\PY{p}{,} \PY{n}{c}\PY{p}{,} \PY{n}{W}\PY{p}{)}
    \PY{n}{h} \PY{o}{=} \PY{n}{bernoulli}\PY{p}{(}\PY{n}{fold\PYZus{}in}\PY{p}{(}\PY{n}{h\PYZus{}key}\PY{p}{,} \PY{n}{step}\PY{p}{)}\PY{p}{,} \PY{n}{nn}\PY{o}{.}\PY{n}{sigmoid}\PY{p}{(}\PY{n}{h\PYZus{}logits}\PY{p}{)}\PY{p}{)}
    \PY{n}{\ensuremath{\rho}} \PY{o}{=} \PY{n}{nn}\PY{o}{.}\PY{n}{sigmoid}\PY{p}{(}\PY{n}{jnp}\PY{o}{.}\PY{n}{einsum}\PY{p}{(}\PY{l+s+s1}{\PYZsq{}}\PY{l+s+s1}{bhc,khc,kcv\PYZhy{}\PYZgt{}bv}\PY{l+s+s1}{\PYZsq{}}\PY{p}{,} \PY{n}{h}\PY{p}{,} \PY{n}{W}\PY{p}{,} \PY{n}{Q}\PY{p}{)}\PY{p}{)}
    \PY{n}{\ensuremath{\nu}} \PY{o}{=} \PY{p}{(}\PY{l+m+mi}{1} \PY{o}{\PYZhy{}} \PY{n}{\ensuremath{\alpha}}\PY{p}{)} \PY{o}{*} \PY{n}{\ensuremath{\nu}} \PY{o}{+} \PY{n}{\ensuremath{\alpha}} \PY{o}{*} \PY{n}{\ensuremath{\rho}} \PY{o}{*} \PY{p}{(}\PY{l+m+mi}{1} \PY{o}{\PYZhy{}} \PY{n}{\ensuremath{\rho}}\PY{p}{)}
    \PY{n}{v} \PY{o}{=} \PY{n}{bernoulli}\PY{p}{(}\PY{n}{fold\PYZus{}in}\PY{p}{(}\PY{n}{v\PYZus{}key}\PY{p}{,} \PY{n}{step}\PY{p}{)}\PY{p}{,} \PY{n}{\ensuremath{\rho}}\PY{p}{)}
    \PY{n}{t}\PY{p}{,} \PY{n}{d} \PY{o}{=} \PY{n}{t} \PY{o}{+} \PY{l+m+mi}{1}\PY{p}{,} \PY{n+nb}{max}\PY{p}{(}\PY{n}{d} \PY{o}{\PYZhy{}} \PY{l+m+mi}{1}\PY{p}{,} \PY{o}{\PYZhy{}}\PY{l+m+mi}{1}\PY{p}{)}
  \PY{k}{return} \PY{n}{s\PYZus{}max}\PY{o}{.}\PY{n}{max}\PY{p}{(}\PY{p}{)}
\end{Verbatim}
    \caption{\small A simplified implementation of RbmSAT; unit propagation and wall clock timing have been stubbed out for clarity. \texttt{rbmsat} takes as arguments canonical OR gate RBM parameters, the table $\mathbf{T}$ (section~\ref{subsec:implementation}), a number of chains $B$ and total number of variables $N$, the unit propagation period \texttt{upp}, the number of steps before fetching unit propagation results \texttt{upw}, a random seed and a parameter $\alpha$ for the moving average of the variance. Returns the maximum number of clauses satisfied when the time budget is exhausted.
    }
    \label{fig:code}
}
\end{figure} 
\subsection{List of problem instances}
\label{subsec:instance_list}
Table~\ref{tab:instance_list} lists the names of the MaxSAT Evaluation instances that are used in this work, split by year. The instances are available for download from the Evaluation websites linked from the top row of the table.

\begin{table*}[t]
    \centering
    \small{
    \begin{tabular}{c c c c}
    \toprule
    \textsc{\href{https://maxsat-evaluations.github.io/2018/}{MaxSAT 2018}} & \textsc{\href{https://maxsat-evaluations.github.io/2018/}{MaxSAT 2019}} & \textsc{\href{https://maxsat-evaluations.github.io/2018/}{MaxSAT 2020}} & \textsc{\href{https://maxsat-evaluations.github.io/2018/}{MaxSAT 2021}}\\
    \textsc{(17 instances)} & \textsc{(29 instances)} & \textsc{(23 instances)} & \textsc{(23 instances)}\\
    \midrule
    \tiny maxcut-hamming6-4.clq.cnf & \tiny scpcyc06\_maxsat.cnf & \tiny ram\_k4\_n20.ra0.cnf & \tiny maxcut-140-630-0.7-6.cnf\\
    \tiny maxcut-brock400\_4.clq.cnf & \tiny scpcyc10\_maxsat.cnf & \tiny normalized-par32-1.cnf & \tiny uaq-min-nr-nr50-np400-rpp5-nc0-rs0-t0-plb100-n9.cnf\\
    \tiny bcp-msp-normalized-par32-1-c.cnf & \tiny MANN\_a27.clq.cnf & \tiny ram\_k4\_n19.ra0.cnf & \tiny ram\_k3\_n18.ra0.cnf\\
    \tiny set-covering-scpcyc07\_maxsat.cnf & \tiny ram\_k4\_n18.ra0.cnf & \tiny MANN\_a9.clq.cnf & \tiny maxcut-140-630-0.7-4.cnf\\
    \tiny bcp-msp-normalized-f600.cnf & \tiny p\_hat700-3.clq.cnf & \tiny sanr400\_0.7.clq.cnf & \tiny maxcut-140-630-0.8-26.cnf\\
    \tiny bcp-msp-normalized-f2000.cnf & \tiny normalized-par32-5.cnf & \tiny MANN\_a81.clq.cnf & \tiny scpcyc09\_maxsat.cnf\\
    \tiny maxcut-brock400\_2.clq.cnf & \tiny normalized-par32-4-c.cnf & \tiny hamming6-2.clq.cnf & \tiny data.243.cnf\\
    \tiny maxcut-t6pm3-8888.spn.cnf & \tiny brock200\_1.clq.cnf & \tiny normalized-g100x100.opb.msat.cnf & \tiny hamming10-4.clq.cnf\\
    \tiny maxcut-brock200\_1.clq.cnf & \tiny sanr200\_0.9.clq.cnf & \tiny ram\_k3\_n19.ra0.cnf & \tiny brock200\_3.clq.cnf\\
    \tiny bcp-msp-normalized-par32-1.cnf & \tiny san200\_0.7\_1.clq.cnf & \tiny scpcyc08\_maxsat.cnf & \tiny ram\_k3\_n14.ra0.cnf\\
    \tiny maxcut-p\_hat500-3.clq.cnf & \tiny p\_hat500-2.clq.cnf & \tiny brock400\_4.clq.cnf & \tiny maxcut-140-630-0.8-18.cnf\\
    \tiny set-covering-scpcyc08\_maxsat.cnf & \tiny ram\_k3\_n16.ra0.cnf & \tiny ram\_k3\_n13.ra0.cnf & \tiny ram\_k3\_n15.ra0.cnf\\
    \tiny set-covering-scpcyc09\_maxsat.cnf & \tiny ram\_k3\_n12.ra0.cnf & \tiny normalized-par32-2.opb.msat.cnf & \tiny maxcut-140-630-0.7-45.cnf\\
    \tiny set-covering-scpcyc06\_maxsat.cnf & \tiny san400\_0.7\_2.clq.cnf & \tiny ram\_k3\_n20.ra0.cnf & \tiny ram\_k3\_n11.ra0.cnf\\
    \tiny set-covering-scpcyc10\_maxsat.cnf & \tiny maxcut-140-630-0.8-7.cnf & \tiny c-fat500-10.clq.cnf &\tiny  MinFill\_R0\_queen5\_5.cnf\\
    \tiny bcp-msp-normalized-f1000.cnf & \tiny ram\_k3\_n10.ra0.cnf & \tiny normalized-par32-3-c.cnf & \tiny maxcut-140-630-0.7-12.cnf\\
    \tiny maxcut-MANN\_a9.clq.cnf & \tiny normalized-par32-1.opb.msat.cnf & \tiny ram\_k3\_n17.ra0.cnf & \tiny data.729.cnf\\
     & \tiny normalized-par32-5.opb.msat.cnf & \tiny scpcyc07\_maxsat.cnf & \tiny data.135.cnf\\
     & \tiny brock400\_3.clq.cnf & \tiny MANN\_a45.clq.cnf & \tiny maxcut-140-630-0.8-39.cnf\\
     & \tiny normalized-par32-4.opb.msat.cnf & \tiny p\_hat500-3.clq.cnf & \tiny maxcut-140-630-0.8-1.cnf\\
     & \tiny keller5.clq.cnf & \tiny johnson8-4-4.clq.cnf & \tiny maxcut-140-630-0.7-20.cnf\\
     & \tiny brock400\_2.clq.cnf & \tiny hamming8-2.clq.cnf & \tiny maxcut-140-630-0.7-28.cnf\\
     & \tiny normalized-par32-3.opb.msat.cnf & \tiny brock400\_1.clq.cnf & \tiny data.405.cnf\\
     & \tiny brock800\_1.clq.cnf &  &  \\
     & \tiny hamming10-2.clq.cnf &  &  \\
     & \tiny maxcut-140-630-0.7-31.cnf &  &  \\
     & \tiny p\_hat300-3.clq.cnf &  &  \\
     & \tiny sanr400\_0.5.clq.cnf &  &  \\
     & \tiny maxcut-140-630-0.8-17.cnf &  &  \\
    \bottomrule
    \end{tabular}
    }
    \caption{\small Names of MaxSAT Evaluation instances used for evaluating RbmSAT.}
    \label{tab:instance_list}
\end{table*}

\begin{table*}[h!]
    \centering
    \begin{sc}
    \begin{tabular}{lccc}
    \toprule
    Evaluation & Prioritized UP Interval (Gibbs steps) & Markov chains per device \\ \midrule
    MaxSAT 2018 & 5000 & 128 \\ MaxSAT 2019 & 5000 & 128 \\ MaxSAT 2020 & 5000 & 128 \\MaxSAT 2021 & 4000 & 64 \\\bottomrule
    \end{tabular}
    \end{sc}
    \caption{\small RbmSAT hyperparameter values used for MaxSAT Evaluations.}
    \label{tab:hyperparams}
\end{table*}

\subsection{Additional implementation details}
\noindent\textbf{Pretraining:}
We pre-train OR-gate RBMs with different satisfying assignment free energy targets for maximum clause sizes of 3 through 7 as described in section \ref{sec:approach}. We use Adam~\cite{kingma2014adam} with a learning rate of $10^{-3}$. We use $3$ hidden units for clause size 3 and $\NumClauseVis + 1$ hidden units for clause sizes $\NumClauseVis \in [4, 7]$.

Our experiments use pre-trained RBMs at either 2, 8 or 16 distinct free energy targets. In the 1 TPU condition in figure~\ref{fig:batchsize_ntpus_scaling}, the two logical cores use OR-gate RBMs trained with $\FreeEnergy^\star \in \{0.068, 0.528\}$. In the 4 TPU condition, targets were 0.068, 0.128, 0.188, 0.248, 0.308, 0.368, 0.428, and 0.488. In the remaining conditions additional intermediate targets were added for a total of 16 targets: 0.068, 0.098, 0.128, 0.158, 0.188, 0.218, 0.248, 0.278, 0.308, 0.338, 0.368, 0.398, 0.428, 0.458, 0.488, and 0.518. For 16 TPUs with 32 logical cores we repeat each distinct set of RBM parameters twice, with distinct random initializations. For 64 TPUs with 128 logical cores, we repeat each distinct set of RBM parameters 4 times.

\noindent\textbf{Precision:}
\label{appendix:precision}
The TPU matrix multiplication unit (MXU) natively performs computations in 16-bit \emph{bfloat16} floating point arithmetic with the same approximate dynamic range as IEEE 754 single precision, employing the same exponent bit width but a reduced resolution significand (8-bit rather than 24-bit).
Single precision floating point can be emulated with multiple passes but by default, 32-bit operands are truncated.
SIMD operations on the Vector Processing Unit (VPU) (such as the addition of $\HidB$ in~\eqref{eq:sparse_h_given_v}) are performed in 32-bit arithmetic for 32-bit operands. 
We store RBM parameters in 32-bit floating point but empirically the speed benefits of 16-bit MXU operation outweigh any envisioned gains from increased precision.

\noindent\textbf{Gather and scatter-add operations:}
We describe a procedure for indexing on the MXU using a one-hot expansion of the index table, $\OneHotExpansion$.
Depending on the instance size, we found performing these computations on the VPU (using JAX ``fancy indexing'' for gathers and \texttt{jax.lax.segment\_sum} for scatter-adds) could result in higher throughput.
Note also that while binary visible unit states are exactly representable in \emph{bfloat16}, the scatter-add computation summing logit contributions will have different precision characteristics when performed on the MXU and VPU, as described in section~\ref{appendix:precision}.
We use this alternative VPU-based implementation in our experimental results.

\noindent\textbf{Batching operations:}
We are able to significantly increase sampling throughput by compiling multiple sampling steps at a time using XLA control flow (\texttt{jax.lax.scan}).
For every dispatch to the TPU, we perform 500 rounds of block Gibbs sampling, where each round consists of sampling hidden unit values and then sampling new visible unit values.

\subsection{Hyperparameters}
\label{subsec:hyperparams}
We tune RbmSAT's hyperparameters for each MaxSAT Evaluation year separately using a grid search.
For a given Evaluation year, each grid search point is evaluated on the instances from all the other Evaluations and the best hyperparameter values are selected.
The selected values are shown in Table~\ref{tab:hyperparams}.

\end{document}